\documentclass{article}
\PassOptionsToPackage{numbers,sort&compress}{natbib}
\usepackage[preprint]{neurips_2026}
\AtBeginDocument{\newgeometry{textwidth=6.5in,textheight=9.3in,top=0.8in,
  headheight=12pt,headsep=20pt,footskip=26pt}}
\usepackage[utf8]{inputenc}
\usepackage[T1]{fontenc}
\usepackage{microtype}
\usepackage{amsmath,amssymb}
\usepackage{graphicx}
\usepackage{subcaption}
\usepackage{booktabs}
\usepackage{multirow}
\usepackage{xcolor}
\usepackage{colortbl}
\usepackage{enumitem}
\usepackage{float}
\definecolor{oursrow}{RGB}{243,248,255}
\newcommand{\oursrow}{\rowcolor{oursrow}}
\usepackage[colorlinks=true,linkcolor=blue,citecolor=blue,urlcolor=blue]{hyperref}
\usepackage[capitalize,noabbrev]{cleveref}
\makeatletter
\renewcommand{\paragraph}{\@startsection{paragraph}{4}{\z@}{0.4ex \@plus 0.2ex \@minus 0.1ex}{-1em}{\normalfont\normalsize\bfseries}}
\makeatother
\newcommand{\hlocal}{Holosoma-Local}
\newcommand{\hglobal}{Holosoma-Global}
\newcommand{\smbase}{Holosoma-SM}
\newcommand{\smours}{\smbase{} [MJWarp]}
\newcommand{\smisaac}{\smbase{} [IsaacSim]}

\title{Holosoma-Tracker: Open, Reproducible, and Deployable Humanoid Whole-body Tracking} 
\author{Jinkun Cao$^{\dagger}$, Shibo Zhao$^{\dagger}$, Adam Setapen, Guanya Shi,\\
  \textbf{Karen Liu, Pieter Abbeel, Rocky Duan$^{\ddagger}$, Carmelo Sferrazza$^{\ddagger}$}\\[4pt] \\
  Amazon FAR (Frontier AI \& Robotics)}
\date{}

\begin{document}
\maketitle
{\renewcommand{\thefootnote}{}\footnotetext{$^{\dagger}$Equal contribution. $^{\ddagger}$Equal advising.}}

\begin{abstract}
Whole-body motion trackers turn human motion into commands for humanoid robots, and have become a core tool
for teleoperation and data collection. In this work, we present Holosoma-Tracker, an open, reproducible, and
deployable whole-body tracker for the Unitree G1. Holosoma-Tracker is open, as it is trained
exclusively on public motion data, about 850 hours from five datasets, and the tracker is released with its code,
checkpoints, and evaluation protocol. It is reproducible, as a single fixed protocol evaluates our trackers
and nine released baselines under the same conditions. It is deployable, as
we evaluate it on the real robot, both under motion capture and in a teleoperation study across operators
and tasks. Its local version, \hlocal{}, supports high-throughput teleoperation with a high success rate on the
tasks that we test. Its global version, \hglobal{}, additionally observes the world pose of the robot, and,
on hardware, reduces the root drift of the local trackers about tenfold with motion capture and fivefold with
lidar-inertial odometry. Finally, the recipe is general, as it trains a tracker not only with PPO, but also with the off-policy
FastSAC and with the world-model-based SuperMimic.

\end{abstract} 

\section{Introduction}
\label{sec:intro}

\begin{figure*}[t]
\centering
\includegraphics[width=\textwidth]{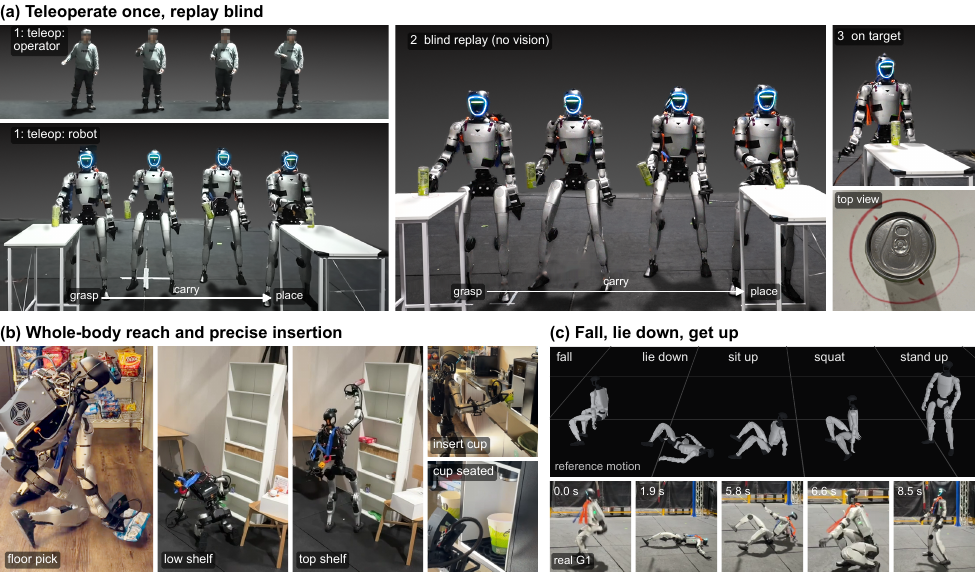}
\caption{\textbf{Holosoma-Tracker on the real Unitree G1.} (a)~A teleoperated pick-and-place (1), replayed by the motion-capture checkpoint of \hglobal{} with lidar-inertial odometry~\cite{superodom} as its only pose source (2), places the bottle on its target (3). (b)~Whole-body reaching, from the floor to the top shelf, and inserting a cup into a coffee machine. (c)~Falling, lying down, and getting up: reference (top) and robot (bottom).}
\label{fig:teaser}
\end{figure*}

Humanoid robots closely resemble humans, which makes human motion a natural source of their behaviors.
Whole-body motion trackers turn human motion from motion capture, video, teleoperation, or a generative
model into control commands for the robot. Trained with large-scale reinforcement learning (RL) on
hundreds of hours of retargeted motion, they follow most everyday and athletic motions in simulation and
on hardware~\cite{sonic,beyondmimic,gmt}, and are a core component of teleoperation and data-collection
systems~\cite{omnih2o,twist,twist2}, which stream the motion of an operator to the robot in real time.

Building on these trackers still poses three challenges. First, reproducing a tracker end to end, and
comparing new ideas on equal footing, requires its training data, code, and evaluation protocol to be
public. Second, trackers are mainly evaluated by tracking errors in simulation, which do not always
reflect their behavior on the robot. Finally, benchmark suites resampled from public motion corpora may
overlap with the training data of trackers that use the same corpora, blurring the line between fitting
and generalization.

We present Holosoma-Tracker, an open and reproducible whole-body tracker for the 29-DoF Unitree G1
(\cref{fig:teaser}), built on Holosoma\footnote{\url{https://github.com/amazon-far/holosoma}}, our
open-source framework for training, deploying, and retargeting humanoid policies. Its local version is
trained exclusively on public motion data, about 175k clips from Bones-seed~\cite{bonesseed}, HiPHI~\cite{hiphi},
AMASS~\cite{amass}, 100STYLE~\cite{style100}, and OMOMO~\cite{omomo}, part of which we retarget with a
pipeline that rejects every physically invalid clip. We design it for loco-manipulation, where it must
track the body and wrists accurately while the robot walks, kneels, and reaches, and remain robust to
payloads and imperfect references. We build on BeyondMimic~\cite{beyondmimic} for
most components and focus on the design choices that serve this goal, measured on hardware wherever
possible.

\begin{figure*}[t]
\centering
\includegraphics[width=\textwidth]{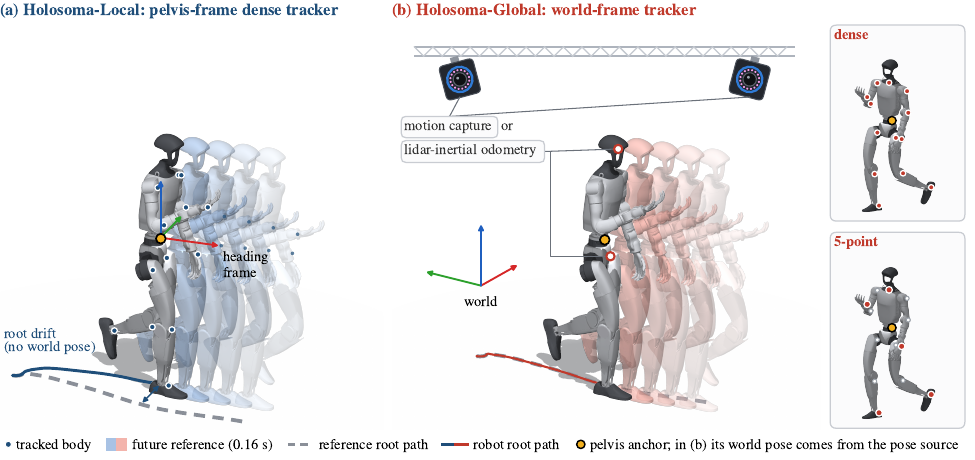}
\caption{The two trackers. (a)~\hlocal{} follows the reference joints and pelvis orientation at the future frames $t{+}2,4,6,8$ (ghosts) in its heading-aligned pelvis frame; dots mark the 14 rewarded bodies (amber: pelvis anchor). Without a world pose, its root can drift. (b)~\hglobal{} also receives the world pose of the robot and follows the reference in the world frame, from a dense or a 5-point (torso, wrists, ankles) command.}
\label{fig:overview}
\end{figure*}

We release two versions (\cref{fig:overview}). \hlocal{} is a local tracker: it follows the full
reference relative to its own body and lets its root drift, which suits teleoperation, where the operator
closes the loop, and a higher-level System-1 (S1) policy that commands the tracker, such as a
vision-language-action model~\cite{groot,sonic}. \hglobal{} is a global tracker: it also observes the reference root
relative to the world pose of the robot, from motion capture or lidar-inertial odometry, which removes most of
the drift. This matters when no operator closes
the loop, as in data pipelines, and when a higher-level policy commands chunks of motion, within which a
local tracker drifts. \hglobal{} is trained with dense and sparse commands, of which the released
inference code runs the dense one. Both versions see only the next 0.16\,s of the reference. A longer lookahead helps
prepare dynamic transitions, but in teleoperation it can only be supplied by delaying the command by the
same amount.

Teleoperation remains the main source of loco-manipulation data, and the tracker bounds its throughput:
it must respond without noticeable lag and must not fail mid-task. Tracking accuracy is often considered
secondary, since the operator corrects errors, but it matters for two reasons. Teleoperated
demonstrations later train policies that command the same tracker, so the recorded commands should match
the executed motion, and precise wrist placement often decides whether a grasp succeeds. We also want a
stack without expensive devices or special infrastructure, so that operators teleoperate challenging
whole-body tasks with a consumer PICO headset over an ordinary network.

We evaluate our trackers under one protocol against nine baselines on held-out motions in simulation, on
hardware under motion capture, and in a teleoperation study across operators and three whole-body
pick-and-place tasks. In simulation, \hlocal{} has lower locomotion body and wrist errors than every
baseline and, on HumanTracker~\cite{humantracker}, completes more clips than the SONIC releases and about as many as
Humanoid-GPT~\cite{humanoidgpt}. Its remaining gaps are highly dynamic and ground motion, outside our design
focus, and long-horizon root drift, which \hglobal{} reduces tenfold on hardware with motion capture.

The recipe is also general. While \hlocal{} trains with PPO~\cite{ppo}, two experimental recipes train the
same tracker with the off-policy FastSAC~\cite{fastsac} and with SuperMimic~\cite{supermimic}, which backpropagates the tracking reward
through a learned world model. Both recipes can considerably reduce the training cost. In fact, in MuJoCo Warp~\cite{mujocowarp}, the SuperMimic recipe
reaches a lower body error than \hlocal{} over the locomotion, manipulation, and dance motions of
MotionDecode~\cite{motiondecode} with less than 1\% of the GPU-hours of \hlocal{} on a single GPU.

Our contributions are listed below:
\begin{itemize}[leftmargin=*]
\item We release Holosoma-Tracker, an open and reproducible whole-body tracker for the G1 that is trained
  exclusively on public motion data, together with the full deployment and evaluation stack, the training code,
  the checkpoints, and a retargeting pipeline with strict physical quality gates.
\item We provide \hlocal{}, a local tracker for teleoperation and for integration with a higher-level S1
  policy, and \hglobal{}, a global tracker that drifts less and, on hardware, tracks more closely.
\item We describe two experimental training recipes, based on off-policy learning and on backpropagation
  through a learned world model, as a starting point for research on faster and cheaper tracker training.
\item We compare our trackers with publicly released trackers in simulation, on hardware under motion capture,
  and in a teleoperation study.
\end{itemize}

\section{Related Works}
\label{sec:related}

\paragraph{Whole-body motion tracking.}
Physics-based tracking began in character animation, where RL policies first imitated a single
clip~\cite{deepmimic} and later most of a large motion corpus~\cite{phc,amass}, or completed partial
goals~\cite{maskedmimic}. On humanoid robots, the sim-to-real
gap became the central problem, addressed with privileged teachers, domain randomization, and filtered motion
data, and extended to object interaction~\cite{h2o,omnih2o,exbody2,hover,gmt,hdmi}. Recent work tunes a single PPO
recipe for highly dynamic motions~\cite{beyondmimic} or scales data and policy toward behavior foundation
models~\cite{sonic,scalebfm,humanoidgpt}. Several trackers are released for the G1~\cite{holomotion,heft,teleopit},
but each trains on its own corpus, which confounds data with recipe. We train \hlocal{} only on public data and
revisit individual design choices, such as the assistive curriculum of ZEST~\cite{zest}. Most trackers follow the
reference relative to their own body, so their root drifts, and world-frame tracking needs the world pose of the robot, from motion capture or from lidar-inertial odometry~\cite{superodom}.

\paragraph{Motion retargeting.}
A tracker is only as good as its references: penetrations or floating contacts in a retargeted clip carry
over to the learned policy, and retargeting quality strongly affects tracking~\cite{gmr}. Most retargeters
solve inverse kinematics on a human--robot body correspondence that is either specified per
robot~\cite{gmr,soma} or learned~\cite{umr}, and interaction-mesh formulations~\cite{intermesh,omniretarget}
also preserve the contacts with objects and terrain. 
We instead treat retargeting as data cleaning and discard every clip that remains physically invalid
(\cref{sec:retarget}).

\paragraph{Learning algorithms.}
PPO~\cite{ppo} is the default training algorithm for most trackers, as it scales to the thousands of parallel environments of GPU-based simulators. The same parallelism has recently made off-policy learning fast
enough for humanoid control~\cite{fasttd3,fastsac}, and SuperMimic~\cite{supermimic} replaces
policy-gradient estimates with gradients through a learned world model, at a fraction of the cost. We test
whether one recipe carries over across these learners by training our tracker with PPO, FastSAC, and
SuperMimic, the latter in MuJoCo Warp~\cite{mujocowarp,mjlab} and in IsaacSim.

\paragraph{Teleoperation.}
Teleoperation systems stream the pose of an operator to a tracker on the
robot~\cite{omnih2o,humanplus,twist,twist2}, or condition the policy on human motion
directly~\cite{anchor1}. Any future reference that a tracker observes must first be produced by the
interface, so its lookahead adds to the command latency. We judge our trackers by the success rate and time per
episode of teleoperated tasks rather than by tracking errors.

\paragraph{Evaluation.}
Trackers are usually reported in their own simulators, robot models, and metrics, which makes published
numbers hard to compare. Shared benchmarks address this by running released policies through one
deployment path on the MotionDecode suites~\cite{sim2real}, by adding human-aligned
metrics~\cite{humantracker}, or by defining open tracking tasks~\cite{motionchallenge}, whereas
HumanoidBench~\cite{humanoidbench} targets task learning. We evaluate on MotionDecode and HumanTracker after
auditing their suites for overlap with our training data, and complement both with measurements on the robot,
which we consider the decisive test. Simulated suites miss how closely a tracker follows the reference under real
sensing and actuation, and in teleoperation a single fall can halt data collection for hours or damage the
hardware. We therefore measure tracking accuracy on the real robot (\cref{sec:exp:sim2real}) and judge
teleoperation by its success rate and throughput, where any fall or stumble fails the trial (\cref{sec:exp:teleop}). 

\section{Method}
\label{sec:method}

\subsection{Problem Formulation}
\label{sec:problem}

\paragraph{Robot and reference motion.}
We consider the 29-joint Unitree G1 humanoid, with 6 joints per leg, 3 in the waist, and 7 per arm. At
50\,Hz, the policy outputs joint-position targets $q^{\ast}_t \in \mathbb{R}^{29}$, which a PD controller
tracks with $\tau = k_p (q^{\ast}_t - q) - k_d \dot q$, clipped at the effort limits,
at the 200\,Hz physics rate in simulation and on the motor drivers on the robot. The policy follows a
reference clip, a 50\,Hz sequence $\{m_k\}_{k=0}^{K}$ of robot poses retargeted from human motion
(\cref{sec:data,sec:retarget}). Each frame holds the root pose, the joint positions and velocities $(\hat
q_k, \dot{\hat q}_k)$, and the world poses and velocities of 30 robot bodies from forward kinematics.
At each step, the policy sees the reference 0.16\,s ahead, its lookahead, which keeps small the
delay with which a teleoperation or streaming source can supply it.

\paragraph{Local and global tracking.}
\hlocal{} follows the reference relative to its own body and never observes its world position, so the
operator or a higher-level policy must correct its drifting root. \hglobal{} also observes the reference
root position relative to its estimated world pose and is rewarded in the world frame
(\cref{tab:hp_global}). Both aim to follow the reference over the whole clip
without falling, which for \hglobal{} includes the root position in the world, and \cref{sec:eval} gives
the failure rule of each benchmark.

\subsection{Model Design}
\label{sec:method:obs}
\label{sec:method:arch}

\paragraph{Proprioception.}
Each frame is 93-D: projected gravity (3), base angular velocity (3), joint positions (29), joint
velocities (29), and the previous action (29). The policy keeps the last 9 frames (0.16\,s) and gathers
five of them, at $\{t, t{-}1, t{-}2, t{-}4, t{-}8\}$, which keeps the recent frames at the full control
rate while covering the whole window.

\paragraph{Reference command.}
The reference enters at the future offsets $\{0,2,4,6,8\}$ at 50\,Hz, a 0.16\,s window sampled every
other step, which halves the command size at the same horizon. For \hlocal{}, each offset holds the
reference joint positions and velocities (29 each) and the orientation of the reference pelvis, our
anchor body, relative to that of the robot (6-D rotation), 320 dimensions in all. This relative
orientation lets the policy follow turns and tilts without observing its world pose. During training, we
add uniform noise to the proprioception and to the anchor orientation of the command.

\paragraph{Critic and actions.}
The privileged critic, used only in training, also observes simulator state: the body poses of the
reference and of the robot, the base linear velocity, and a 10-frame history of joint states and actions.
The policy outputs offsets $a_t$ around a default standing pose with a per-joint scale of
$0.25\,\tau^{\max}_j / k_{p,j}$, so that a unit action corresponds to a quarter of the effort limit at
zero velocity error; actions and PD torques are clipped.

\begin{figure}[t]
\centering
\includegraphics[width=0.66\textwidth]{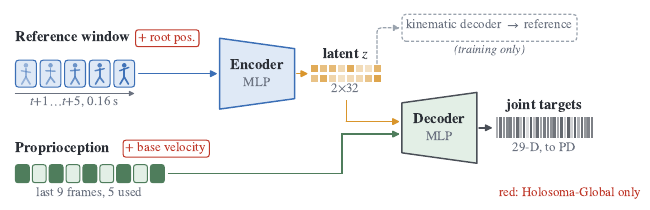}
\caption{Encoder--decoder actor of Holosoma-Tracker. Inputs in red are used only by
\hglobal{}, which replaces the MLPs with transformer blocks. The reference window holds the five reference
frames of the command (\cref{sec:method:obs}), labeled $t{+}1$ to $t{+}5$.}
\label{fig:arch}
\end{figure}

\paragraph{Policy architecture.}
Both trackers use an encoder--decoder actor (\cref{fig:arch}). In \hlocal{}, an MLP \emph{command
encoder} compresses the reference window into a continuous 64-D latent, and an MLP \emph{action decoder}
maps this latent and the gathered history to the mean action. The latent separates what to track from
how to act: the decoder sees the reference only through this code, so other commands can drive the same decoder through their own encoders. \hglobal{} adds such an encoder for its sparse commands, with a latent tied to the dense one, and replaces the MLPs of the actor with transformer blocks, while the critic remains an MLP. A scaled and shifted $\tanh$ bounds each latent element to a fixed range. An auxiliary \emph{kinematic decoder} reconstructs the command from the latent, which regularizes the latent toward a motion code.

\subsection{Training}
\label{sec:method:training}
\label{sec:method:reward}
\label{sec:method:dr}

\paragraph{Rewards and terminations.}
The reward is a weighted sum of Gaussian kernels and regularizers. Its tracking terms
compare 14 bodies: the pelvis, the torso, and on each side the hip, knee, ankle, shoulder, elbow, and
wrist. For \hlocal{}, body poses are compared relative to the pelvis of each pose, with yaw and
horizontal position removed. Only the two anchor terms and the world-frame velocity terms see the world
frame, so drift is penalized only weakly, until the world-frame anchor-position kernel saturates. An
end-effector term on both wrists and on a point 0.5\,m above the torso sharpens upper-body tracking, and
the penalty on undesired contacts skips the contacts that the reference itself makes with the floor. An
episode ends with its clip or when the anchor, ankle, wrist, or foot error exceeds a threshold.

\paragraph{Robot model.}
Training, evaluation, and deployment use the standard model-15 G1 URDF without payload,
with the torque and velocity limits and the armature of each joint. We train with the implicit position control
of the simulator and without the self-collisions between the pelvis and the hip-roll links, which overlap
in deep hip flexion.

\paragraph{Randomization.}
Each environment draws its friction, restitution, base center of mass, torso and wrist masses, and joint
calibration bias once at startup, and redraws its initial joint state and push schedule (one push every
4--6\,s) at every reset. Both trackers train on slightly uneven ground. In 70\%
of the episodes, the reference of the actor carries a slowly drifting pitch bias that models the
pelvis-pitch drift of a head-mounted pose estimate.

\paragraph{Optimization.}
We use PPO~\cite{ppo} with a KL-adaptive learning rate. \hlocal{} trains
262{,}144 environments on 64 L40S GPUs for 40k iterations.
Each GPU trains on an active pool of 1{,}024 clips, sampled in 50-frame bins in proportion to their
failure rates. Low-posture clips, mostly from the rarest family
of our corpus, enter the pool with ten times their weight (the low-posture weight, \cref{tab:hp_sampler}).
Each GPU holds only a small part of the corpus, so its pool must turn over. We \emph{roll} the pool:
each iteration replaces the four or five clips that have stayed longest, which turns the whole pool over
every 250 iterations, and each environment switches clips at its own reset.

\subsection{Global Tracking}
\label{sec:method:global}

\paragraph{Command.}
At each offset, the dense command also holds the reference root position relative to the current root
position of the robot, in its pelvis frame, and each proprioceptive frame also holds the base linear
velocity.
Training also samples sparse modes, whose command holds only the torso, wrists, and ankles (5-point), the
torso and wrists, or only the wrists. Each keypoint is the reference pose of one body relative to the
current pose of the same body of the robot, and the sparse encoder of \cref{sec:method:arch} processes the
sparse command. We evaluate the dense and 5-point modes in \cref{tab:main_md,tab:ht}.

\paragraph{Control and training.}
\hglobal{} uses explicit torque control, which computes the PD torque outside the physics solver at every
physics step and clips it at the effort limit, instead of the implicit position drive of \hlocal{}. We train it in two stages. The first stage trains a
motion-capture tracker from scratch, with the ground-truth root pose and a $4\times$ smaller batch than
\hlocal{}, and the second fine-tunes it to be robust to the noise of lidar-inertial odometry.

\paragraph{Deployment modes.}
\hglobal{} takes its orientation from the IMU, and root position
and base linear velocity from motion capture or Super Odometry~\cite{superodom} on the robot and from the
true root pose in simulation. The lidar-inertial odometry checkpoint, which we run with Super
Odometry, also receives a pose-validity flag and is trained with the pose withheld in about half
of the samples, so that it keeps tracking as a local tracker without a pose.

\section{Motion Data}
\label{sec:data}

\paragraph{Training corpus.}
\hlocal{} trains on about 850 hours of public motion data, about 175k G1 clips at 50\,Hz
including left--right mirrors, from five datasets. Most clips come from
Bones-seed~\cite{bonesseed}, 288 hours of public motion capture, whose G1 retargets we use after the
robot filter of the public SONIC training pipeline~\cite{sonic}, which drops motions that need props
absent in simulation and some acrobatics. HiPHI~\cite{hiphi}, released only as human
motion and retargeted by us, provides more than 60\% of the hours with long, high-precision captures of
human motion and object interaction. G1 retargets of AMASS~\cite{amass}, which unifies many earlier
motion-capture datasets, add a broad range of everyday motion, and those of 100STYLE~\cite{style100} and
OMOMO~\cite{omomo} add stylized locomotion and object interaction.

\paragraph{Retargeting.}
\label{sec:retarget}
Our retargeting pipeline (\cref{supp:ret}) combines the collision-aware IK of GMR~\cite{gmr} with the
body correspondence and fallback solver of SoMA~\cite{soma}, then smooths the joint trajectories, repairs
short discontinuities, and lifts the root out of floor penetrations. To avoid trading physical validity
for yield, it discards every clip that self-collides, penetrates the floor, has floating contacts,
violates the joint limits, or moves implausibly. Of the 35{,}252 HiPHI clips (17{,}626 sequences and their
left--right mirrors), it keeps 30{,}652, which amount to 532 hours of training data.

\section{Evaluation Protocol}
\label{sec:eval}

Every policy, including the baselines, runs in MuJoCo~\cite{mujoco} through the same open-source
deployment harness~\cite{sim2real}, under a protocol that we fix before any comparison
(\cref{supp:eval}). As low tracking errors do not guarantee deployability, we also evaluate deployment on the
real robot (\cref{sec:exp:sim2real,sec:exp:teleop}).

\paragraph{Suites.}
MotionDecode~\cite{motiondecode} has four suites named by their clip counts: we pool Locomotion-80,
Manipulation-48, and Dance-40 into MD-168 and report Ground-60 separately, as ground motion is outside
the focus of our trackers. HumanTracker~\cite{humantracker} has a test split of 2{,}500 clips, which we
run with its own simulation setup, robot model, and failure criterion. No evaluated clip is in
our training data.

\paragraph{Metrics.}
On MotionDecode, a rollout fails at the first frame with a pelvis-height error above 0.30\,m. Besides the
success rate, we report the body and wrist position errors (mm) over the frames before failure, in the local
frame, which is centered on the pelvis and aligned with its heading, and the global root error. The hardware experiments
also report the heading and wrist orientation errors (deg), and the payload study reports the progress, i.e., the
completed fraction of each clip. As a policy that falls early skips the often hardest frames, we always
report errors with success or progress. HumanTracker scores success rate, joint-position error,
and a learned human-likeness score.

\paragraph{Baselines.}
We compare with nine released G1 trackers on both benchmarks: three SONIC releases~\cite{sonic} (the main release, which is our reference baseline, the low-latency
SONIC-LL, and v1.1) and six MotionDecode
leaderboard checkpoints~\cite{sim2real}, including the latest of Humanoid-GPT~\cite{humanoidgpt}. For
each baseline with published results, we reproduce the results reported on the benchmarks
on which it was originally evaluated.

\section{Simulation Experiments}
\label{sec:experiments}
\label{sec:exp}

Following the protocol of \cref{sec:eval}, we compare our trackers with the baselines
(\cref{sec:exp:main}), then report the experimental training recipes (\cref{sec:exp:alt}); the appendix reports the ablations
(\cref{sec:exp:ablations}) and the behavior relevant for deployment
(\cref{sec:exp:deploy,sec:exp:payload,sec:exp:gait}).

\paragraph{Models.}
We report \hlocal{} after 40k iterations and \hglobal{} with the true root pose and with its dense and
5-point commands. Of the experimental recipes (\cref{sec:exp:alt}), we report \smbase{} in MuJoCo Warp
(MJWarp)~\cite{mjlab,mujocowarp} and in IsaacSim after 100k iterations, as its base recipe varies
little across seeds, although its Ground-60 success varies widely (\cref{fig:sm_optim}), and
Holosoma-FastSAC after 100k iterations.

\subsection{Main Results}
\label{sec:exp:main}

\begin{table*}[t]
\centering\footnotesize
\caption{\textbf{Results on the held-out MotionDecode suites.} Local-frame body and wrist position errors (mm), Locomotion end root error (m), and success rate (\%; a rollout fails when its pelvis-height error exceeds 0.30\,m). Look.: command lookahead (s). Bold: best per column. $^\S$Reference observation from a substitute (TeleopIT) or pinned (HoloMotion) G1 model.}
\label{tab:main_md}
\setlength{\tabcolsep}{2.8pt}
\begin{tabular}{lcccccccccccccc}
\toprule
& & \multicolumn{4}{c}{Locomotion-80} & \multicolumn{3}{c}{Manipulation-48} & \multicolumn{3}{c}{Ground-60} &
\multicolumn{3}{c}{Dance-40} \\
\cmidrule(lr){3-6}\cmidrule(lr){7-9}\cmidrule(lr){10-12}\cmidrule(lr){13-15}
Method & Look.\ (s) & Body & Wrist & Root & Succ & Body & Wrist & Succ & Body & Wrist & Succ & Body & Wrist & Succ \\
\midrule
SONIC~\cite{sonic} & 0.90 & 29.73 & 30.61 & 0.527 & 100.0 & 45.62 & 56.78 & 100.0 & 78.2 & 89.6 & 66.7 & 48.52 & 63.46 & 97.5 \\
SONIC-LL~\cite{sonic} & 0.18 & 30.96 & 33.37 & 1.077 & 100.0 & 38.97 & 44.02 & 100.0 & 114.1 & 132.3 & 25.0 & 53.52 & 72.34 & 85.0 \\
SONIC v1.1~\cite{sonic} & 0.90 & 30.42 & 32.47 & 0.691 & 100.0 & 40.71 & 44.85 & 100.0 & 124.3 & 140.9 & 20.0 & 52.88 & 70.72 & 95.0 \\
HEFT~\cite{heft} & 0.12 & 31.88 & 30.03 & 0.561 & 100.0 & 36.03 & 38.17 & 97.9 & 98.5 & 105.4 & 56.7 & 44.52 & 49.10 & 87.5 \\
Humanoid-GPT~\cite{humanoidgpt} & 0.02 & 38.55 & 46.52 & 1.626 & 95.0 & 35.67 & 41.41 & 97.9 & 144.5 & 165.8 & 0.0 & 46.66 & 71.40 & 87.5 \\
TeleopIT~\cite{teleopit}$^\S$ & 0.00 & 37.49 & 47.39 & 0.459 & 100.0 & 36.25 & 46.36 & 100.0 & 139.9 & 160.8 & 70.0 & 58.83 & 84.43 & 92.5 \\
ScaleBFM-M~\cite{scalebfm} & 0.10 & 36.87 & 39.10 & 1.101 & 100.0 & 36.05 & 35.66 & 100.0 & 98.9 & 112.7 & 80.0 & 49.70 & 54.93 & 97.5 \\
ScaleBFM-XL~\cite{scalebfm} & 0.10 & 38.66 & 54.87 & 0.766 & 100.0 & 37.76 & 48.18 & 100.0 & 89.4 & 107.9 & 85.0 & 48.06 & 60.02 & \textbf{100.0} \\
HoloMotion~\cite{holomotion}$^\S$ & 0.20 & 33.01 & 30.24 & 1.166 & 100.0 & 43.01 & 41.14 & 95.8 & 109.2 & 131.7 & 10.0 & \textbf{40.13} & 47.68 & 90.0 \\
\oursrow \begin{tabular}[t]{@{}l@{}}\smbase{}~\cite{supermimic} [MJWarp]\end{tabular} & 0.16 & \textbf{24.94} & \textbf{24.29} & 0.537 & 100.0 & \textbf{24.38} & \textbf{19.36} & 100.0 & 61.8 & 60.9 & 93.3 & 41.38 & \textbf{44.03} & \textbf{100.0} \\
\oursrow \begin{tabular}[t]{@{}l@{}}\smbase{}~\cite{supermimic} [IsaacSim]\end{tabular} & 0.16 & 28.18 & 28.17 & 0.649 & 100.0 & 30.92 & 21.31 & 100.0 & 72.0 & 71.1 & 95.0 & 45.91 & 46.98 & 95.0 \\
\oursrow Holosoma-FastSAC~\cite{fastsac} & 0.16 & 36.71 & 37.69 & 2.581 & 93.8 & 41.34 & 52.60 & 97.9 & 149.1 & 174.3 & 0.0 & 49.33 & 61.63 & 37.5 \\
\oursrow \hlocal{} & 0.16 & 26.81 & 27.48 & 0.749 & 100.0 & 33.85 & 41.88 & 100.0 & 75.1 & 82.8 & 78.3 & 43.39 & 52.93 & 95.0 \\
\oursrow \hglobal{} & 0.16 & 27.88 & 33.24 & \textbf{0.030} & 100.0 & 28.54 & 32.41 & 100.0 & 69.2 & 80.1 & 86.7 & 43.17 & 55.69 & 92.5 \\
\oursrow \hglobal{}, 5-point & 0.16 & 31.20 & 35.74 & 0.034 & 100.0 & 32.25 & 36.18 & 100.0 & 97.7 & 103.4 & \textbf{96.7} & 47.09 & 59.04 & 97.5 \\
\bottomrule
\end{tabular}
\end{table*}

\paragraph{MotionDecode.}
\hlocal{} has lower Locomotion body and wrist errors than every baseline, and tracks the body more closely
than every baseline on Manipulation and than all but HoloMotion on Dance. Some baselines remain ahead on the
Manipulation wrist error and on the Locomotion root drift, which \hlocal{} does not observe by design, and on Ground-60, mostly getting up and moving on the floor, only the two ScaleBFM models
complete more clips. With the true root pose, \hglobal{} reduces the Locomotion root error of \hlocal{} by more
than an order of magnitude, and tracks the body and wrists on Manipulation more closely and completes more
Ground-60 clips than every baseline, at the cost of higher Locomotion body and wrist errors than \hlocal{}. Its 5-point command
keeps the root almost as close, with moderately higher body errors.

\begin{table*}[t]
\centering\footnotesize
\setlength{\tabcolsep}{2pt}
\caption{\textbf{Results on the HumanTracker test split} (2{,}500 clips) by motion family: success (\%), MPJPE (rad), and human-likeness score (HS, 0--100). Bold: best per column. $^\S$See \cref{tab:main_md}.}
\label{tab:ht}
\begin{tabular*}{\linewidth}{@{}l@{\extracolsep{\fill}\hspace{6pt}}ccccccccccccccc@{}}
\toprule
 & \multicolumn{3}{c}{Daily} & \multicolumn{3}{c}{Highly Dynamic} & \multicolumn{3}{c}{Interaction} & \multicolumn{3}{c}{Ground} & \multicolumn{3}{c}{Overall} \\
\cmidrule(lr){2-4}\cmidrule(lr){5-7}\cmidrule(lr){8-10}\cmidrule(lr){11-13}\cmidrule(lr){14-16}
Method & Succ & MPJPE & HS & Succ & MPJPE & HS & Succ & MPJPE & HS & Succ & MPJPE & HS & Succ & MPJPE & HS \\
\midrule
SONIC~\cite{sonic} & 93.6 & 0.103 & 49.5 & 84.0 & 0.118 & 40.8 & 97.5 & 0.126 & 54.9 & 19.5 & 0.232 & 27.1 & 89.4 & 0.116 & 50.2 \\
SONIC-LL~\cite{sonic} & 93.4 & 0.105 & 33.2 & 84.3 & 0.124 & 34.8 & 96.3 & 0.130 & 45.2 & 25.6 & 0.233 & 26.5 & 89.2 & 0.119 & 37.3 \\
SONIC v1.1~\cite{sonic} & 94.1 & 0.093 & 43.7 & 84.7 & 0.108 & 38.6 & \textbf{97.9} & 0.109 & 48.8 & 13.4 & 0.198 & 34.1 & 89.5 & 0.103 & 44.9 \\
HEFT~\cite{heft} & 95.0 & 0.050 & 44.3 & 78.0 & 0.057 & 40.4 & 97.2 & 0.060 & 50.8 & 15.9 & 0.176 & 36.0 & 88.9 & \textbf{0.058} & 46.1 \\
Humanoid-GPT~\cite{humanoidgpt} & 94.3 & \textbf{0.046} & \textbf{54.8} & \textbf{87.3} & \textbf{0.047} & \textbf{49.0} & 97.2 & 0.069 & 56.8 & \textbf{32.3} & 0.215 & 25.1 & \textbf{90.7} & 0.059 & \textbf{54.2} \\
TeleopIT~\cite{teleopit}$^\S$ & 85.3 & 0.097 & 33.4 & 58.6 & 0.110 & 36.1 & 96.5 & 0.105 & 46.2 & 12.8 & 0.229 & 47.4 & 82.6 & 0.104 & 38.5 \\
ScaleBFM-M~\cite{scalebfm} & 92.7 & 0.092 & 43.7 & 72.0 & 0.091 & 41.6 & 96.9 & 0.084 & 57.9 & 17.7 & 0.187 & 59.8 & 87.4 & 0.092 & 49.1 \\
ScaleBFM-XL~\cite{scalebfm} & 92.3 & 0.092 & 38.3 & 75.7 & 0.089 & 36.8 & 97.0 & 0.082 & 51.4 & 17.1 & 0.159 & 51.7 & 87.6 & 0.090 & 43.2 \\
HoloMotion~\cite{holomotion}$^\S$ & 92.2 & 0.075 & 32.0 & 67.2 & 0.086 & 36.9 & 96.9 & 0.084 & 48.3 & 9.8 & 0.171 & 45.0 & 86.2 & 0.082 & 38.4 \\
\oursrow \begin{tabular}[t]{@{}l@{}}\smbase{}~\cite{supermimic} [MJWarp]\end{tabular} & 89.2 & 0.056 & 38.6 & 55.6 & 0.061 & 34.8 & 96.6 & \textbf{0.056} & 46.4 & 28.7 & 0.133 & 32.8 & 84.9 & \textbf{0.058} & 40.9 \\
\oursrow \begin{tabular}[t]{@{}l@{}}\smbase{}~\cite{supermimic} [IsaacSim]\end{tabular} & 86.3 & 0.060 & 27.0 & 44.8 & 0.067 & 29.7 & 95.7 & 0.063 & 42.9 & 22.0 & 0.160 & 30.7 & 81.8 & 0.065 & 32.9 \\
\oursrow Holosoma-FastSAC~\cite{fastsac} & 81.7 & 0.138 & 36.1 & 49.6 & 0.207 & 33.6 & 94.4 & 0.163 & 49.6 & \phantom{0}4.3 & 0.352 & 31.9 & 78.8 & 0.158 & 40.6 \\
\oursrow \hlocal{} & \textbf{96.3} & 0.067 & 48.0 & 84.7 & 0.082 & 43.7 & 97.0 & 0.076 & \textbf{59.4} & 18.9 & 0.196 & 32.8 & 90.3 & 0.075 & 51.2 \\
\oursrow \hglobal{} & 87.1 & 0.073 & 23.1 & 53.4 & 0.095 & 28.1 & 96.2 & 0.073 & 43.0 & 18.3 & 0.172 & 21.7 & 82.9 & 0.077 & 30.3 \\
\oursrow \hglobal{}, 5-point & 87.2 & 0.127 & 15.7 & 54.1 & 0.148 & 27.1 & 95.8 & 0.127 & 37.3 & 17.7 & 0.224 & 24.9 & 82.8 & 0.131 & 24.3 \\
\bottomrule
\end{tabular*}
\end{table*}

\paragraph{HumanTracker.}
\label{sec:exp:ht}
We run every tracker in the harness, robot model, and termination rule of the benchmark
(\cref{sec:eval}), each baseline with its own observation, lookahead, and gains, and
\hglobal{} with the true root state. 
\hlocal{} has the highest success rate on the Daily family, and overall it completes more clips than the
SONIC releases and about as many as Humanoid-GPT, within the checkpoint-to-checkpoint variation (\cref{supp:noise}), with a lower
joint error than every SONIC release. On the Highly Dynamic family, only Humanoid-GPT completes more clips. HEFT
and Humanoid-GPT have the lowest joint errors of the baselines, Humanoid-GPT has the highest human-likeness
score, followed by \hlocal{}, SONIC, and ScaleBFM-M, and two baselines complete more clips of the Ground family, while SONIC completes about as many. \hglobal{} completes fewer clips than every baseline except TeleopIT, as it completes
fewer of the Daily clips and only about half of the Highly Dynamic ones, failing most of those in which
the reference raises a foot high, such as kicks, and both of its modes have lower human-likeness scores
than every baseline. This gap may come from its $4\times$ smaller batch, which lowers the HumanTracker success of the base
recipe (\cref{sec:exp:ablations}), and from less than a quarter of the environment steps of \hlocal{}
(\cref{sec:method:global,tab:compute}).

\begin{figure*}[t]
\centering
\includegraphics[width=\textwidth]{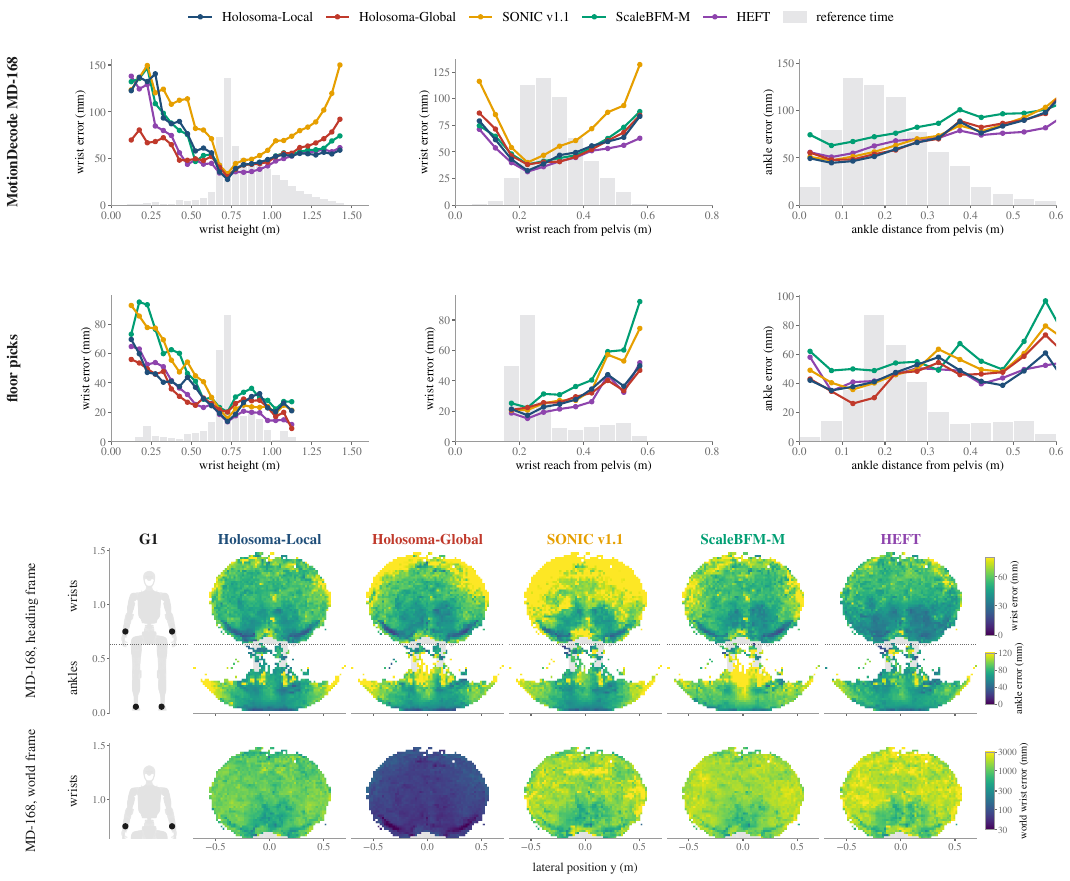}
\caption{Where each tracker loses accuracy (simulation). Top: wrist error against wrist height and reach, and ankle error against ankle distance from the pelvis, on MD-168 and on eight Bones-seed floor picks. Bottom: MD-168 error over the front-view workspace for the wrists and ankles (local frame) and for the wrists (world frame, aligned by the initial root pose only), beside the G1 at the same scale.}
\label{fig:ws_trackers}
\end{figure*}

\paragraph{Where trackers lose accuracy.}
\Cref{fig:ws_trackers} breaks the wrist and ankle errors down by target position relative to the pelvis.
The five trackers it compares all reach nearly the whole reference workspace, so they differ in error,
not in reach, and each is most accurate where the references spend most of their time, a region that
therefore dominates the means of \cref{tab:main_md}. Only \hglobal{} keeps the wrists within a few
centimeters of the reference in the world frame (\cref{supp:workspace}).

\subsection{Experimental Training Recipes}
\label{sec:exp:alt}

Two experimental recipes train the tracker of \cref{sec:method} with other learners, the off-policy
FastSAC~\cite{fastsac} and the world-model-based SuperMimic~\cite{supermimic}. Both keep the corpus and command
interface of \hlocal{}, and, as far as each learner allows, its reward and domain randomization. FastSAC keeps all
of them unchanged but changes the networks, while SuperMimic keeps the networks but uses its own termination
(\cref{supp:sm,supp:fastsac}).

\paragraph{SuperMimic.}
Beyond the learner and the termination, \smbase{} replaces the episode-start sampler, trains on flat ground, and
drops the clip augmentations. It keeps the three additions of \cref{sec:exp:ablations}, i.e., the reference-contact
rule, the low-posture weight, and the pelvis--hip-roll filter, which the MuJoCo model of the robot already applies. As in
SuperMimic, it terminates on pelvis height and tilt and passes every randomized physical quantity (friction,
joint offsets, torso center of mass, mass scales) to both the world model and the critic.
On a single GPU, with less than 1\% of the GPU-hours of \hlocal{}, it matches \hlocal{}'s MD-168 body error in
IsaacSim and beats it in MuJoCo Warp. In MuJoCo Warp it also has the lowest Locomotion and Manipulation body and
wrist errors of all trackers. In both simulators it completes more Ground-60 clips than \hlocal{} and drifts less
on Locomotion. The IsaacSim policy has higher body errors than the MuJoCo Warp policy on every suite
(\cref{tab:main_md}), i.e., it transfers less well to our MuJoCo evaluation.
On HumanTracker, its joint error is below \hlocal{}'s, and in MuJoCo Warp it is the lowest of all trackers, tied
with HEFT. However, it completes fewer clips than \hlocal{} and scores lower on
human-likeness (\cref{tab:ht}), so low joint error before a failure does not predict completion. Under the
stricter termination of \hlocal{}, its base recipe falls on long clips and drifts much further on Dance-40 in both
simulators (\cref{supp:sm}).

\paragraph{FastSAC.}
FastSAC~\cite{fastsac}, an off-policy variant of SAC, learns two distributional critics from a replay
buffer and updates the actor on the mean of their Q-values. We adapt its original configuration for
whole-body tracking and otherwise train \hlocal{}'s recipe unchanged. The
recipe thus carries over to an off-policy learner, although Holosoma-FastSAC trails \hlocal{} in tracking error
and success, and its root drifts further on Locomotion. 

\section{Real-World Experiments}
\label{sec:exp:real}

\subsection{Reference Motion Tracking}
\label{sec:exp:sim2real}

\paragraph{Setup.}
In the motion-capture setup, OptiTrack cameras track a set of markers on the pelvis of the robot at 200\,Hz, and the pose reaches the
policy over ROS~2 with a median latency of 2--6\,ms. Super Odometry~\cite{superodom} runs on a laptop
next to the robot and estimates the pose of the head-mounted Livox MID-360 lidar at 200\,Hz. The policy
always runs onboard at 50\,Hz (\cref{supp:eval}).
Every controller runs onboard through the same deployment stack, on four clips of 13.5--32\,s: turning with small steps,
walking forward, backward, and sideways, a crouched walk, and deep squats with single-leg balance. Each
reads the clip through its own command interface, with a 0.16\,s lookahead for our trackers and 0.9\,s
for SONIC v1.1, from the same start pose, to which we align the reference (\cref{sec:method:global}). 

\begin{table*}[t]
\centering\footnotesize
\caption{\textbf{Reference motion tracking on the real G1}, averaged over four clips: root error over the clip and at its end (m), heading error (deg), local-frame body and wrist position (mm) and wrist orientation (deg) errors, and step count and foot lift relative to the reference (1.00 matches it). Bold: best per column, or closest to 1.00. }
\label{tab:sim2real}
\setlength{\tabcolsep}{4pt}
\begin{tabular}{lcccccccc}
\toprule
 & \multicolumn{3}{c}{Global} & \multicolumn{3}{c}{Pose} & \multicolumn{2}{c}{Gait} \\
\cmidrule(lr){2-4}\cmidrule(lr){5-7}\cmidrule(lr){8-9}
Controller & Root, mean & Root, end & Heading & Body & Wrist & Wrist ori. & Steps $\times$ref & Lift $\times$ref \\
\midrule
SONIC v1.1~\cite{sonic}             & 0.65 & 1.22 & 2.5  & 29.4 & \textbf{32.6} & 8.2  & 1.22 & 0.72 \\
\oursrow \hlocal{}                  & 0.59 & 1.15 & 2.5  & 28.7 & 33.0  & 11.9 & \textbf{0.94} & 0.72 \\
\oursrow \hglobal{}, Super Odometry & 0.13 & 0.12 & 2.6 & 31.8 & 36.0 & 6.1 & 1.50 & 1.13 \\
\oursrow \hglobal{}, motion capture, 5-point & 0.12 & 0.14 & 3.5 & 40.9 & 48.0 & 6.3 & 2.22 & 1.07 \\
\oursrow \hglobal{}, motion capture & \textbf{0.06} & \textbf{0.06} & \textbf{2.4} & \textbf{27.1} & 32.9 & \textbf{5.1} & 1.30 & \textbf{1.04} \\
\bottomrule
\end{tabular}
\end{table*}

\paragraph{Results.}
With motion capture, \hglobal{} keeps its root close to the reference over the whole clip, whereas the
roots of \hlocal{} and SONIC v1.1 drift about an order of magnitude further (\cref{tab:sim2real}). Their end error is about twice the mean, so their drift keeps growing, whereas for
\hglobal{} the two are about equal, as the position feedback removes drift as it builds up and leaves an
error set by the pose source. \hglobal{} also has the lowest body and wrist-orientation errors of all
controllers and nearly the lowest wrist error. With the lidar-inertial odometry checkpoint and Super Odometry, the root error doubles but stays
about five times lower than that of the local trackers, and with the 5-point command, its root stays
about as close as with Super Odometry, at a higher body error.

\paragraph{Sim-to-real gap.}
Replaying every hardware run in simulation with the same deployment code and settings shows that
simulation predicts the body pose well but not the root drift of our trackers: on hardware, the root of \hlocal{}
drifts about as far as that of SONIC v1.1 (\cref{tab:sim2real}), although on these clips it drifts much less in simulation.
Position feedback makes the root insensitive to this gap: with motion capture, the root error of \hglobal{} stays
within a few centimeters.

\subsection{Teleoperation}
\label{sec:exp:teleop}

\begin{figure*}[t]
\centering
\resizebox{\textwidth}{!}{\includegraphics[height=2.4in]{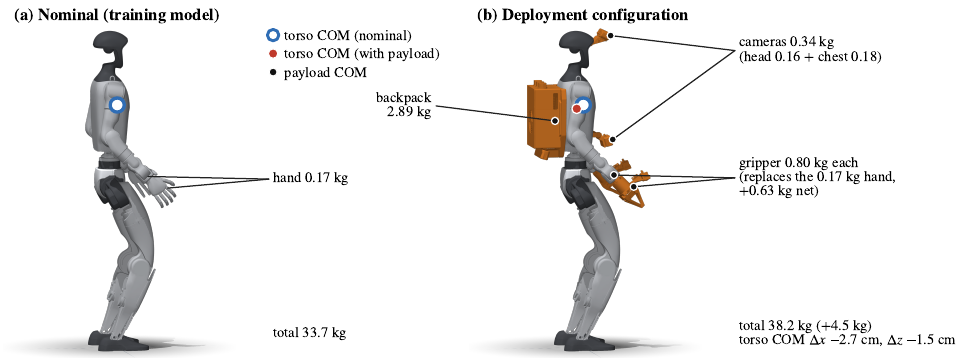}\hspace{0.12in}%
\includegraphics[height=2.4in]{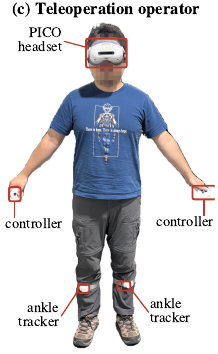}}
\caption{(a)~The nominal model-15 G1 used in training (33.7\,kg). (b)~The teleoperation configuration with backpack, cameras, and grippers (+4.5\,kg; payload in orange, centers of mass in black, torso center of mass blue to red). (c)~The operator's PICO headset, hand controllers, and ankle trackers.}
\label{fig:payload_configs}
\end{figure*}

\paragraph{Online retargeting.}
In every trial, a consumer PICO headset (\cref{fig:payload_configs}(c))
estimates a 24-joint body pose of the operator from the headset, two hand controllers, and two ankle
trackers. A GMR retargeter~\cite{gmr} solves the inverse kinematics of each pose and streams the
resulting G1 joint positions and root pose to the robot as a 50\,Hz reference.

\paragraph{Tasks.}
We compare the trackers by their success rate and time per episode on three whole-body pick-and-place
(PnP) tasks. In \textbf{Kneel-down PnP}, the robot kneels to pick a can from
the floor and places it into a box on a table. In \textbf{Bi-manual PnP}, it sorts four objects into two
bins, one hand per bin. In \textbf{4-in-a-row PnP}, it places four cans onto stacks of increasing height.
The placement of the objects changes between trials, and every trial uses the configuration of
\cref{fig:payload_configs}(b), whose payload no tracker is trained with. \Cref{supp:teleop} shows the tasks and details their
procedure, randomization, and failure conditions.

\paragraph{Protocol.}
Each task is performed by four operators in two shifts. Before data collection, each operator trains with every tested tracker for two shifts, and the operators are not told which tracker is running. A trial succeeds if none of the failure conditions occurs. Otherwise, the operator ends it
at the failure, and its time still counts toward the mean episode time over all trials ($t_\text{all}$).

\paragraph{Results.}
\hlocal{} achieves the highest success rate on all three tasks (\cref{tab:teleop}). Against the SONIC
releases in G1-mode, which receive the same retargeted reference, the gap is largest on Kneel-down PnP,
which requires the most whole-body motion. \hlocal{} also completes Kneel-down and Bi-manual PnP faster
than both, while SONIC v1.1 is faster on 4-in-a-row PnP. G1-mode is not the teleoperation interface of
SONIC, since a live stream cannot supply its 0.9\,s reference window, which we compress to 0.36\,s on our
stations. SONIC v1.1 in SMPL-mode, which tracks the body pose of the operator directly, also succeeds
less often than \hlocal{} on every task, with about the same episode times on Kneel-down and Bi-manual
PnP.

\begin{table*}[t]
\centering\footnotesize
\caption{\textbf{Teleoperation results on the real G1}, pooled over operators: success rate (\%) and mean episode time over all ($t_\text{all}$) and successful ($t_\text{succ}$) trials (s). SONIC G1-mode reads the retargeted robot joints; SMPL-mode reads the body pose of the operator. }
\label{tab:teleop}
\setlength{\tabcolsep}{4pt}
\begin{tabular}{lccccccccc}
\toprule
& \multicolumn{3}{c}{Kneel-down PnP} & \multicolumn{3}{c}{Bi-manual PnP} & \multicolumn{3}{c}{4-in-a-row PnP} \\
\cmidrule(lr){2-4}\cmidrule(lr){5-7}\cmidrule(lr){8-10}
Controller & Succ. & $t_\text{all}$ & $t_\text{succ}$ & Succ. & $t_\text{all}$ & $t_\text{succ}$ & Succ. & $t_\text{all}$ & $t_\text{succ}$ \\
\midrule
SONIC~\cite{sonic} [G1-mode] & 12.2 & 34.4 & 33.0 & 73.2 & 51.1 & 55.9 & 71.1 & 63.5 & 70.6 \\
SONIC v1.1~\cite{sonic} [G1-mode] & 10.5 & 35.8 & 28.9 & 76.2 & 48.1 & 54.6 & 73.4 & 52.4 & 59.4\\
SONIC v1.1~\cite{sonic} [SMPL-mode] & 71.3 & \textbf{20.9}  & \textbf{20.8} & 67.3 & \textbf{36.1} & 41.4 & 82.5 & \textbf{38.0} & \textbf{40.1} \\
\oursrow \hlocal{} & \textbf{72.6} & 21.3 & 22.2 & \textbf{77.2} & 37.6 & \textbf{39.8} & \textbf{85.2} & 60.7 & 66.2 \\
\bottomrule
\end{tabular}
\end{table*}

\section{Limitations}
\label{sec:limitations}

\paragraph{Online retargeting.}
In teleoperation, the accuracy of trackers remains bounded by the online retargeting, whose errors and latency reach the tracker as an imperfect reference. Tracking the body pose of
the operator directly would avoid this step, but for many tasks we still find it advantageous to decouple
retargeting from tracking, as this records an explicit pair of reference and executed G1 motions for every teleoperated demonstration.

\paragraph{Pose estimation.}
\hglobal{} removes most of the drift, but relies on the accuracy of its pose estimate. With the lidar-inertial
odometry checkpoint and Super Odometry, the root error on hardware is about twice that with motion capture (\cref{sec:exp:sim2real}), and more
accurate onboard pose estimation would directly improve global tracking without external infrastructure.

\paragraph{Evaluation.}
As the baselines are trained on other corpora, our comparisons do not separate the contribution of the
training data from that of the training recipe. On hardware, we compare with a single baseline, SONIC v1.1,
on four clips (\cref{sec:exp:sim2real}).
These results show large effects, such as the drift that \hglobal{} removes, but small differences between
controllers may not hold over more runs.

\paragraph{Training cost.}
\hlocal{} and \hglobal{} are trained with PPO, and \hlocal{} requires 6{,}000 L40S GPU-hours. The SuperMimic
recipe of \cref{sec:exp:alt} reduces this cost by more than two orders of magnitude, but remains
experimental and needs further study before it can replace PPO.

\section{Conclusion}
\label{sec:conclusion}

We presented Holosoma-Tracker, an open, reproducible, and deployable whole-body tracker for the Unitree
G1, released with its training, deployment, retargeting, and evaluation code and checkpoints, whose local
version trains only on public motion data. \hlocal{} tracks the body and wrists during locomotion in simulation more closely than all nine
baselines, \hglobal{} removes most of the
root drift on hardware, and the same reward and nearly the same randomization also train a SuperMimic tracker with less
than 1\% of the GPU-hours. As simulation errors alone can mislead, we report them together with success
rates and measure the trackers on the robot, and hope that Holosoma-Tracker serves as a strong and
reproducible baseline for humanoid whole-body tracking.

\bibliographystyle{plainnat}
\bibliography{refs}

\begin{thebibliography}{49}
\providecommand{\natexlab}[1]{#1}
\providecommand{\url}[1]{\texttt{#1}}
\expandafter\ifx\csname urlstyle\endcsname\relax
  \providecommand{\doi}[1]{doi: #1}\else
  \providecommand{\doi}{doi: \begingroup \urlstyle{rm}\Url}\fi

\bibitem[Araujo et~al.(2025)Araujo, Ze, Xu, Wu, and Liu]{gmr}
Joao~Pedro Araujo, Yanjie Ze, Pei Xu, Jiajun Wu, and C.~Karen Liu.
\newblock Retargeting matters: General motion retargeting for humanoid motion
  tracking.
\newblock \emph{arXiv preprint arXiv:2510.02252}, 2025.

\bibitem[{Archon Robotics}(2026)]{anchor1}
{Archon Robotics}.
\newblock Anchor 1: Precise and stable humanoid control for whole-body
  manipulation.
\newblock \url{https://archon.tech/research/anchor1}, 2026.

\bibitem[{Bones Studio}(2026)]{bonesseed}
{Bones Studio}.
\newblock {BONES-SEED} motion dataset.
\newblock \url{https://bones.studio/}, 2026.
\newblock Motion Data by Bones Studio.

\bibitem[Cao et~al.(2026)Cao, Fang, Kwon, Yu, Ma, Tan, Zhou, Du, Gu, Gao, Dai,
  Han, and Xu]{umr}
Hanyang Cao, Yuetong Fang, Taesoo Kwon, Runyi Yu, Ji~Ma, Jing Tan, Yangchen
  Zhou, Baoze Du, Yi~Gu, Yukang Gao, Ruoli Dai, Lei Han, and Renjing Xu.
\newblock Unified motion retargeting for humanoids with learned point cloud
  correspondence.
\newblock \emph{arXiv preprint arXiv:2609.02134}, 2026.

\bibitem[Casiez et~al.(2012)Casiez, Roussel, and Vogel]{oneeuro}
G{\'e}ry Casiez, Nicolas Roussel, and Daniel Vogel.
\newblock 1{\texteuro} filter: A simple speed-based low-pass filter for noisy
  input in interactive systems.
\newblock In \emph{ACM CHI Conference on Human Factors in Computing Systems},
  2012.

\bibitem[Chen et~al.(2026)Chen, Wang, Zhang, Ma, Yang, Ren, Huang, Zhu, Wang,
  and Su]{holomotion}
Maiyue Chen, Kaihui Wang, Bo~Zhang, Xihan Ma, Zhiyuan Yang, Yi~Ren, Qijun
  Huang, Zihao Zhu, Yucheng Wang, and Zhizhong Su.
\newblock {HoloMotion-1} technical report.
\newblock \emph{arXiv preprint arXiv:2605.15336}, 2026.

\bibitem[Chen et~al.(2025)Chen, Ji, Cheng, Peng, Peng, and Wang]{gmt}
Zixuan Chen, Mazeyu Ji, Xuxin Cheng, Xuanbin Peng, Xue~Bin Peng, and Xiaolong
  Wang.
\newblock {GMT}: General motion tracking for humanoid whole-body control.
\newblock \emph{arXiv preprint arXiv:2506.14770}, 2025.

\bibitem[{Chingmu}(2026)]{motiondecode}
{Chingmu}.
\newblock {MotionDecode}.
\newblock \url{https://huggingface.co/datasets/CMRobot/MotionDecode}, 2026.
\newblock Motion data from Chingmu.

\bibitem[Fu et~al.(2024)Fu, Zhao, Wu, Wetzstein, and Finn]{humanplus}
Zipeng Fu, Qingqing Zhao, Qi~Wu, Gordon Wetzstein, and Chelsea Finn.
\newblock {HumanPlus}: Humanoid shadowing and imitation from humans.
\newblock In \emph{Conference on Robot Learning (CoRL)}, 2024.

\bibitem[{Google DeepMind}(2025)]{mujocowarp}
{Google DeepMind}.
\newblock {MuJoCo Warp}: {GPU}-optimized {MuJoCo} for {NVIDIA} hardware.
\newblock \url{https://github.com/google-deepmind/mujoco_warp}, 2025.

\bibitem[Harvey et~al.(2020)Harvey, Yurick, Nowrouzezahrai, and Pal]{lafan}
F{\'e}lix~G. Harvey, Mike Yurick, Derek Nowrouzezahrai, and Christopher Pal.
\newblock Robust motion in-betweening.
\newblock \emph{ACM Transactions on Graphics (SIGGRAPH)}, 2020.

\bibitem[He et~al.(2024{\natexlab{a}})He, Luo, He, Xiao, Zhang, Zhang, Kitani,
  Liu, and Shi]{omnih2o}
Tairan He, Zhengyi Luo, Xialin He, Wenli Xiao, Chong Zhang, Weinan Zhang, Kris
  Kitani, Changliu Liu, and Guanya Shi.
\newblock {OmniH2O}: Universal and dexterous human-to-humanoid whole-body
  teleoperation and learning.
\newblock In \emph{Conference on Robot Learning (CoRL)}, 2024{\natexlab{a}}.

\bibitem[He et~al.(2024{\natexlab{b}})He, Luo, Xiao, Zhang, Kitani, Liu, and
  Shi]{h2o}
Tairan He, Zhengyi Luo, Wenli Xiao, Chong Zhang, Kris Kitani, Changliu Liu, and
  Guanya Shi.
\newblock Learning human-to-humanoid real-time whole-body teleoperation.
\newblock In \emph{IEEE/RSJ International Conference on Intelligent Robots and
  Systems (IROS)}, 2024{\natexlab{b}}.

\bibitem[He et~al.(2025)He, Xiao, Lin, Luo, Xu, Jiang, Kautz, Liu, Shi, Wang,
  Fan, and Zhu]{hover}
Tairan He, Wenli Xiao, Toru Lin, Zhengyi Luo, Zhenjia Xu, Zhenyu Jiang, Jan
  Kautz, Changliu Liu, Guanya Shi, Xiaolong Wang, Linxi Fan, and Yuke Zhu.
\newblock {HOVER}: Versatile neural whole-body controller for humanoid robots.
\newblock In \emph{ICRA}, 2025.

\bibitem[Ho et~al.(2010)Ho, Komura, and Tai]{intermesh}
Edmond S.~L. Ho, Taku Komura, and Chiew-Lan Tai.
\newblock Spatial relationship preserving character motion adaptation.
\newblock \emph{ACM Transactions on Graphics (SIGGRAPH)}, 2010.

\bibitem[Ji et~al.(2026)Ji, Ma, Zhang, Yu, Wang, Chi, Peng, Yan, Gu, Tian, Wu,
  Li, Yuan, Dai, and Han]{hiphi}
Jiahao Ji, Ji~Ma, Runhan Zhang, Runyi Yu, Wenjia Wang, Weiheng Chi, Qianqian
  Peng, Weichao Yan, Yongfei Gu, Ye~Tian, Ting Wu, Longwei Li, Chun Yuan, Ruoli
  Dai, and Lei Han.
\newblock {HiPHI}: A large-scale benchmark for high-precision human motion and
  object-interaction.
\newblock \emph{arXiv preprint arXiv:2608.16222}, 2026.

\bibitem[Ji et~al.(2024)Ji, Peng, Liu, Li, Yang, Cheng, and Wang]{exbody2}
Mazeyu Ji, Xuanbin Peng, Fangchen Liu, Jialong Li, Ge~Yang, Xuxin Cheng, and
  Xiaolong Wang.
\newblock {ExBody2}: Advanced expressive humanoid whole-body control.
\newblock \emph{arXiv preprint arXiv:2412.13196}, 2024.

\bibitem[Jordan et~al.(2024)Jordan, Jin, Boza, You, Cesista, Newhouse, and
  Bernstein]{muon}
Keller Jordan, Yuchen Jin, Vlado Boza, Jiacheng You, Franz Cesista, Laker
  Newhouse, and Jeremy Bernstein.
\newblock Muon: An optimizer for hidden layers in neural networks.
\newblock \url{https://kellerjordan.github.io/posts/muon/}, 2024.

\bibitem[Li et~al.(2023)Li, Wu, and Liu]{omomo}
Jiaman Li, Jiajun Wu, and C.~Karen Liu.
\newblock Object motion guided human motion synthesis.
\newblock \emph{ACM Transactions on Graphics (SIGGRAPH Asia)}, 2023.

\bibitem[Liao et~al.(2025)Liao, Truong, Huang, Gao, Tevet, Sreenath, and
  Liu]{beyondmimic}
Qiayuan Liao, Takara~E. Truong, Xiaoyu Huang, Yuman Gao, Guy Tevet, Koushil
  Sreenath, and C.~Karen Liu.
\newblock {BeyondMimic}: From motion tracking to versatile humanoid control via
  guided diffusion.
\newblock \emph{arXiv preprint arXiv:2508.08241}, 2025.

\bibitem[Liao et~al.(2026)Liao, Shao, Long, Zakka, Horowitz, Abbeel, and
  Sreenath]{supermimic}
Qiayuan Liao, Yiyang Shao, Junfeng Long, Kevin Zakka, Roberto Horowitz, Pieter
  Abbeel, and Koushil Sreenath.
\newblock {SuperMimic}: Scaling humanoid whole-body control with one
  differentiable world model.
\newblock \emph{arXiv preprint arXiv:2610.28480}, 2026.

\bibitem[Liu et~al.(2026{\natexlab{a}})Liu, Lu, Yang, Shi, Yang, Guo, and
  Chen]{heft}
Chenxin Liu, Qingzhou Lu, Guangxiao Yang, Xuanyang Shi, Chenghan Yang, Yanjiang
  Guo, and Jianyu Chen.
\newblock {HEFT}: Heavy-payload full-size humanoid teleoperation with
  privileged motion guidance and windowed payload curriculum.
\newblock \emph{arXiv preprint arXiv:2607.02332}, 2026{\natexlab{a}}.

\bibitem[Liu et~al.(2026{\natexlab{b}})Liu, Qi, Zeng, Yu, Guan, Zhang, Chen,
  Liang, Li, Lin, Yu, Zhang, Wang, and Yi]{humantracker}
Dairu Liu, Zekun Qi, Jiayu Zeng, Ruixi Yu, Yu~Guan, Yintianrun Zhang, Xuchuan
  Chen, Sikai Liang, Zekai Li, Chenghuai Lin, Xinqiang Yu, Wenyao Zhang,
  He~Wang, and Li~Yi.
\newblock {HumanTracker}: Towards comprehensive and human-aligned motion
  tracking benchmark.
\newblock \emph{arXiv preprint arXiv:2608.13555}, 2026{\natexlab{b}}.

\bibitem[Luo et~al.(2023)Luo, Cao, Winkler, Kitani, and Xu]{phc}
Zhengyi Luo, Jinkun Cao, Alexander Winkler, Kris Kitani, and Weipeng Xu.
\newblock Perpetual humanoid control for real-time simulated avatars.
\newblock In \emph{ICCV}, 2023.

\bibitem[Luo et~al.(2026)Luo, Yuan, Wang, Li, Casta{\~n}eda, Chen, Cao, Li,
  Minor, Ben, Park, Sami, Wang, Da, Ding, Hogg, Song, Lim, Jeong, He, Xue,
  Xiao, Yuen, Kautz, Chang, Iqbal, Fan, and Zhu]{sonic}
Zhengyi Luo, Ye~Yuan, Tingwu Wang, Chenran Li, Fernando Casta{\~n}eda, Sirui
  Chen, Zi-Ang Cao, Jiefeng Li, David Minor, Qingwei Ben, Jinhyung Park, David
  Sami, Zi~Wang, Xingye Da, Runyu Ding, Cyrus Hogg, Lina Song, Edy Lim, Eugene
  Jeong, Tairan He, Haoru Xue, Wenli Xiao, Simon Yuen, Jan Kautz, Yan Chang,
  Umar Iqbal, Linxi Fan, and Yuke Zhu.
\newblock {SONIC}: Supersizing motion tracking for natural humanoid whole-body
  control.
\newblock \emph{Science Robotics}, 11\penalty0 (117):\penalty0 eaed4592, 2026.
\newblock arXiv:2511.07820.

\bibitem[Mahmood et~al.(2019)Mahmood, Ghorbani, Troje, Pons-Moll, and
  Black]{amass}
Naureen Mahmood, Nima Ghorbani, Nikolaus~F. Troje, Gerard Pons-Moll, and
  Michael~J. Black.
\newblock {AMASS}: Archive of motion capture as surface shapes.
\newblock In \emph{ICCV}, 2019.

\bibitem[Mason et~al.(2022)Mason, Starke, and Komura]{style100}
Ian Mason, Sebastian Starke, and Taku Komura.
\newblock Real-time style modelling of human locomotion via feature-wise
  transformations and local motion phases.
\newblock \emph{Proc.\ ACM Comput.\ Graph.\ Interact.\ Tech.}, 2022.

\bibitem[{MotionChallenge Team}(2026)]{motionchallenge}
{MotionChallenge Team}.
\newblock {MotionChallenge}: An open, reproducible benchmark for humanoid
  motion tracking.
\newblock \url{https://motionchallenge-project.github.io/motionchallenge/},
  2026.

\bibitem[{NVIDIA}(2026)]{soma}
{NVIDIA}.
\newblock {SOMA} retargeter.
\newblock \url{https://github.com/NVIDIA/soma-retargeter}, 2026.

\bibitem[{NVIDIA} et~al.(2025){NVIDIA}, Bjorck, Casta{\~n}eda, Cherniadev, Da,
  Ding, Fan, Fang, Fox, Hu, Huang, et~al.]{groot}
{NVIDIA}, Johan Bjorck, Fernando Casta{\~n}eda, Nikita Cherniadev, Xingye Da,
  Runyu Ding, Linxi Fan, Yu~Fang, Dieter Fox, Fengyuan Hu, Spencer Huang,
  et~al.
\newblock {GR00T N1}: An open foundation model for generalist humanoid robots.
\newblock \emph{arXiv preprint arXiv:2503.14734}, 2025.

\bibitem[Peng et~al.(2018)Peng, Abbeel, Levine, and van~de Panne]{deepmimic}
Xue~Bin Peng, Pieter Abbeel, Sergey Levine, and Michiel van~de Panne.
\newblock {DeepMimic}: Example-guided deep reinforcement learning of
  physics-based character skills.
\newblock \emph{ACM Transactions on Graphics}, 2018.

\bibitem[Qi et~al.(2026)Qi, Chen, Liu, Lin, Lian, Liang, Zhang, Guan, Wang,
  Zhang, et~al.]{humanoidgpt}
Zekun Qi, Xuchuan Chen, Dairu Liu, Chenghuai Lin, Yunrui Lian, Sikai Liang,
  Zhikai Zhang, Yu~Guan, Jilong Wang, Wenyao Zhang, et~al.
\newblock {Humanoid-GPT}: Scaling data and structure for zero-shot motion
  tracking.
\newblock \emph{arXiv preprint arXiv:2606.03985}, 2026.

\bibitem[{RoboParty Lab Team}(2026)]{sim2real}
{RoboParty Lab Team}.
\newblock sim2real: A lightweight and modular sim2sim and sim2real deployment
  stack.
\newblock \url{https://github.com/EGalahad/sim2real}, 2026.

\bibitem[Schulman et~al.(2017)Schulman, Wolski, Dhariwal, Radford, and
  Klimov]{ppo}
John Schulman, Filip Wolski, Prafulla Dhariwal, Alec Radford, and Oleg Klimov.
\newblock Proximal policy optimization algorithms.
\newblock \emph{arXiv preprint arXiv:1707.06347}, 2017.

\bibitem[Seo et~al.(2025{\natexlab{a}})Seo, Sferrazza, Chen, Shi, Duan, and
  Abbeel]{fastsac}
Younggyo Seo, Carmelo Sferrazza, Juyue Chen, Guanya Shi, Rocky Duan, and Pieter
  Abbeel.
\newblock Learning sim-to-real humanoid locomotion in 15 minutes.
\newblock \emph{arXiv preprint arXiv:2512.01996}, 2025{\natexlab{a}}.

\bibitem[Seo et~al.(2025{\natexlab{b}})Seo, Sferrazza, Geng, Nauman, Yin, and
  Abbeel]{fasttd3}
Younggyo Seo, Carmelo Sferrazza, Haoran Geng, Michal Nauman, Zhao-Heng Yin, and
  Pieter Abbeel.
\newblock {FastTD3}: Simple, fast, and capable reinforcement learning for
  humanoid control.
\newblock \emph{arXiv preprint arXiv:2505.22642}, 2025{\natexlab{b}}.

\bibitem[Sferrazza et~al.(2024)Sferrazza, Huang, Lin, Lee, and
  Abbeel]{humanoidbench}
Carmelo Sferrazza, Dun-Ming Huang, Xingyu Lin, Youngwoon Lee, and Pieter
  Abbeel.
\newblock {HumanoidBench}: Simulated humanoid benchmark for whole-body
  locomotion and manipulation.
\newblock \emph{arXiv preprint arXiv:2403.10506}, 2024.

\bibitem[Sleiman et~al.(2026)Sleiman, Li, Adu-Bredu, Deits, Kumar, Bergamin,
  Bhardwaj, Biddlestone, Burger, Estrada, Iacobelli, Koolen, Lambert, Lin,
  Mungai, Nobles, Rozen-Levy, Shi, Wang, Welner, Yu, Zhang, Rizzi, Hodgins,
  Bertrand, Abe, Kuindersma, and Farshidian]{zest}
Jean~Pierre Sleiman, He~Li, Alphonsus Adu-Bredu, Robin Deits, Arun Kumar, Kevin
  Bergamin, Mohak Bhardwaj, Scott Biddlestone, Nicola Burger, Matthew~A.
  Estrada, Francesco Iacobelli, Twan Koolen, Alexander Lambert, Erica Lin,
  M.~Eva Mungai, Zach Nobles, Shane Rozen-Levy, Yuyao Shi, Jiashun Wang, Jakob
  Welner, Fangzhou Yu, Mike Zhang, Alfred Rizzi, Jessica Hodgins, Sylvain
  Bertrand, Yeuhi Abe, Scott Kuindersma, and Farbod Farshidian.
\newblock {ZEST}: Zero-shot embodied skill transfer for athletic robot control.
\newblock \emph{arXiv preprint arXiv:2602.00401}, 2026.

\bibitem[Tessler et~al.(2024)Tessler, Guo, Nabati, Chechik, and
  Peng]{maskedmimic}
Chen Tessler, Yunrong Guo, Ofir Nabati, Gal Chechik, and Xue~Bin Peng.
\newblock {MaskedMimic}: Unified physics-based character control through masked
  motion inpainting.
\newblock \emph{ACM Transactions on Graphics (SIGGRAPH Asia)}, 2024.

\bibitem[Todorov et~al.(2012)Todorov, Erez, and Tassa]{mujoco}
Emanuel Todorov, Tom Erez, and Yuval Tassa.
\newblock {MuJoCo}: A physics engine for model-based control.
\newblock In \emph{IEEE/RSJ International Conference on Intelligent Robots and
  Systems (IROS)}, 2012.

\bibitem[Weng et~al.(2025)Weng, Li, Sobanbabu, Wang, Luo, He, Ramanan, and
  Shi]{hdmi}
Haoyang Weng, Yitang Li, Nikhil Sobanbabu, Zihan Wang, Zhengyi Luo, Tairan He,
  Deva Ramanan, and Guanya Shi.
\newblock {HDMI}: Learning interactive humanoid whole-body control from human
  videos.
\newblock \emph{arXiv preprint arXiv:2509.16757}, 2025.

\bibitem[Wu et~al.(2026)Wu, Xu, Fan, Li, and Huang]{teleopit}
Bingqian Wu, Zicheng Xu, Xianghui Fan, Dayu Li, and Xiangru Huang.
\newblock {TeleopIT}: A full-embodiment humanoid teleoperation system.
\newblock \emph{arXiv preprint arXiv:2608.01834}, 2026.

\bibitem[Yang et~al.(2025)Yang, Huang, Wu, Kanazawa, Abbeel, Sferrazza, Liu,
  Duan, and Shi]{omniretarget}
Lujie Yang, Xiaoyu Huang, Zhen Wu, Angjoo Kanazawa, Pieter Abbeel, Carmelo
  Sferrazza, C.~Karen Liu, Rocky Duan, and Guanya Shi.
\newblock {OmniRetarget}: Interaction-preserving data generation for humanoid
  whole-body loco-manipulation and scene interaction.
\newblock \emph{arXiv preprint arXiv:2509.26633}, 2025.

\bibitem[Zakka(2024)]{mink}
Kevin Zakka.
\newblock Mink: Python inverse kinematics based on {MuJoCo}.
\newblock \url{https://github.com/kevinzakka/mink}, 2024.

\bibitem[Zakka et~al.(2026)Zakka, Liao, Yi, Le~Lay, Sreenath, and
  Abbeel]{mjlab}
Kevin Zakka, Qiayuan Liao, Brent Yi, Louis Le~Lay, Koushil Sreenath, and Pieter
  Abbeel.
\newblock mjlab: A lightweight framework for {GPU}-accelerated robot learning.
\newblock \emph{arXiv preprint arXiv:2601.22074}, 2026.

\bibitem[Ze et~al.(2025{\natexlab{a}})Ze, Chen, Ara{\'u}jo, Cao, Peng, Wu, and
  Liu]{twist}
Yanjie Ze, Zixuan Chen, Jo{\~a}o~Pedro Ara{\'u}jo, Zi-ang Cao, Xue~Bin Peng,
  Jiajun Wu, and C.~Karen Liu.
\newblock {TWIST}: Teleoperated whole-body imitation system.
\newblock In \emph{Conference on Robot Learning (CoRL)}, 2025{\natexlab{a}}.

\bibitem[Ze et~al.(2025{\natexlab{b}})Ze, Zhao, Wang, Kanazawa, Duan, Abbeel,
  Shi, Wu, and Liu]{twist2}
Yanjie Ze, Siheng Zhao, Weizhuo Wang, Angjoo Kanazawa, Rocky Duan, Pieter
  Abbeel, Guanya Shi, Jiajun Wu, and C.~Karen Liu.
\newblock {TWIST2}: Scalable, portable, and holistic humanoid data collection
  system.
\newblock \emph{arXiv preprint arXiv:2511.02832}, 2025{\natexlab{b}}.

\bibitem[Zeng et~al.(2026)Zeng, Yin, Niu, Lu, Zhong, Chen, Jia, Chen, Wang, Xu,
  et~al.]{scalebfm}
Weishuai Zeng, Kangning Yin, Xiaojie Niu, Shunlin Lu, Weixiang Zhong, Jiahe
  Chen, Feiyu Jia, Xiao Chen, Zirui Wang, Furui Xu, et~al.
\newblock Scaling behavior foundation model for humanoid robots.
\newblock \emph{arXiv preprint arXiv:2607.15163}, 2026.

\bibitem[Zhao et~al.(2021)Zhao, Zhang, Wang, Nogueira, and Scherer]{superodom}
Shibo Zhao, Hengrui Zhang, Peng Wang, Lucas Nogueira, and Sebastian Scherer.
\newblock {Super Odometry}: {IMU}-centric {LiDAR}-visual-inertial estimator for
  challenging environments.
\newblock \emph{arXiv preprint arXiv:2104.14938}, 2021.

\end{thebibliography}

\appendix
\raggedbottom\setlength{\textfloatsep}{20pt plus 2pt minus 4pt}\setlength{\dbltextfloatsep}{20pt plus 2pt minus 4pt}
\renewcommand{\thetable}{S\arabic{table}}\setcounter{table}{0}
\renewcommand{\thefigure}{S\arabic{figure}}\setcounter{figure}{0}
\renewcommand{\theHtable}{S\arabic{table}}\renewcommand{\theHfigure}{S\arabic{figure}}
\clearpage
\begin{center}{\Large\bfseries Appendix}\end{center}
\section{Evaluation Protocol Details}
\label[appendix]{supp:eval}

\paragraph{Protocol and execution.}
Before any comparison, we fix the suites, metrics, averaging, reference baseline, and harness version of
\cref{sec:eval}, and every evaluation checks that the reference baseline reproduces its
recorded results. We report MotionDecode and HumanTracker separately and play each clip once, in full, at 50\,Hz,
on MotionDecode with our G1 model of training and deployment (\cref{sec:method:dr}) and on HumanTracker with its
own robot model (\cref{supp:ht}). The HumanTracker test split~\cite{humantracker} covers daily, highly dynamic,
interaction, and ground motions.

\paragraph{Held-out check.}
For the held-out rule of \cref{sec:eval}, we search each training dataset for clips, source takes, and content
(matched by aligning the body positions of whole clips) shared with each suite. The new captures of HumanTracker
can only match by content, and do so only for steady walking, in 126 clips (100 with
Bones-seed, 34 with AMASS, 8 with both). As AMASS predates HumanTracker yet matches as closely, these are generic
walking matches. Without these clips and their mirrors, no tracker loses more than 0.7 points of overall success
or changes rank, so we keep the full test split. MotionDecode shares no clips or source takes with our corpus, but
13 Locomotion clips contain a 3.5--17\,s walking or jogging segment that matches by content. Without them,
\hlocal{} keeps lower Locomotion body and wrist errors than every baseline and a wider lead over the main SONIC
release, so we keep the full suite.

\paragraph{Bodies, failure, and metrics.}
Tracking errors use the 14 harness bodies (the pelvis, the torso, and on each side the hip-yaw, knee, toe,
shoulder-yaw, elbow, and wrist links), with the robot and reference poses each in their own local frame, centered on the pelvis and aligned with its heading (\cref{sec:eval}). At the hips,
feet, and shoulders, our reward (\cref{sec:method:reward}) uses the hip-, ankle-, and shoulder-roll links instead.
Beyond the failure rule of MotionDecode (\cref{sec:eval}), a rollout on training data also fails at a pelvis-height
error above 0.25\,m or a projected-gravity (torso-tilt) error above 0.8. The success rate is the fraction of clips
completed without failure, and position errors average over bodies and the frames before failure. The global root
error compares the root displacement from the start with that of the reference, at the end of each Locomotion
clip (\cref{tab:main_md}). Metrics weight clips equally.

\paragraph{Checkpoint-to-checkpoint variation.}
\label[appendix]{supp:noise}
As our rollouts are deterministic, differences within the checkpoint-to-checkpoint variation, the spread of a
metric over four late checkpoints of a run of the development recipe (48.5k--50k iterations, \cref{supp:ablations}), count as ties:
\begin{itemize}
\item MotionDecode body-position error: 0.7\,mm on Locomotion-80 and Dance-40, 1.4\,mm on Manipulation-48,
11.8\,mm on Ground-60, and 0.57\,mm on MD-168; MD-168 wrist-position error: 0.48\,mm.
\item HumanTracker success rate: 0.6 points; on the clips with a valid joint error at all four checkpoints,
joint-position error 0.002\,rad and human-likeness score 3.4 points (2.0 points when each checkpoint is scored on its
own valid clips).
\end{itemize}
Two runs of the same recipe with different code versions and seeds, this development run and an earlier one, differ
more, by 2.7\,mm on Locomotion-80 and 3.1\,mm on Dance-40.

\paragraph{Deployment and hardware.}
The torque demand of \cref{sec:exp:deploy} is the median over the 12 walking clips of the largest 99th percentile,
on each clip, of the PD torque $|k_p(q^*-q)-k_d\dot q|$ over the ankle- and waist-pitch joints, relative to their
35\,Nm limit; the maximum is always at an ankle-pitch joint. On hardware, every controller runs our inference runtime on the
Unitree SDK, with ONNX Runtime on the robot CPU. Under motion capture, we estimate the root linear velocity by
finite differences, calibrate the marker-to-pelvis offset once, and align the motion-capture yaw with the
IMU at the start of each run (\cref{sec:exp:sim2real}). With Super Odometry, we map the estimated lidar pose to the
pelvis through the lidar mount and waist joints.

\section{Hyperparameters}
\label[appendix]{app:hparams}

\Cref{tab:hp_ppo,tab:hp_sampler} list the PPO hyperparameters and the clip sampler and data augmentation of
\hlocal{}, and \cref{tab:hp_global} lists the differences of \hglobal{}. The networks of \hlocal{} are SiLU MLPs:
the command encoder has hidden widths (2048, 1024, 512, 512) and a 64-D latent, the action decoder and the critic
(2048, 2048, 1024, 1024, 512, 512), and the auxiliary kinematic decoder (2048, 1024, 512, 512), with its
reconstruction loss weighted by 0.01.

\begin{table}[H]
\centering\footnotesize
\caption{\textbf{PPO hyperparameters of \hlocal{}.} \Cref{tab:hp_global} lists the differences for \hglobal{}.}
\label{tab:hp_ppo}
\begin{tabular}{@{}lp{0.58\linewidth}@{}}
\toprule
Optimizer & AdamW, shared actor and critic \\
Learning rate & $2\times10^{-5}$, KL-adaptive (target 0.01), bounds $[10^{-5}, 2\times10^{-4}]$ \\
Clip / $\gamma$ / $\lambda$ & 0.2 / 0.99 / 0.95 \\
Value coef.\ / entropy / grad-norm clip & 1.0 / 0.01 / 0.1 \\
Epochs / minibatches / steps per env & 5 / 4 / 24 \\
Environments & 262{,}144 on 64 L40S (ablations: the same, except the batch ablation and the development rows of \cref{tab:ablate}) \\
Iterations & 40{,}000 \\
Initial / minimum action std & 0.05 / 0.001; maximum 0.5 \\
Observation normalization & critic only: running mean/variance, clipped to $\pm$5 (actor: raw) \\
Episode length & until the clip ends (no time-based timeout) \\
Simulation / control rate & 200\,Hz / 50\,Hz \\
\bottomrule
\end{tabular}
\end{table}

\begin{table}[H]
\centering\footnotesize
\caption{\textbf{Adaptive clip sampler and data augmentation.}}
\label{tab:hp_sampler}
\begin{tabular}{@{}lp{0.58\linewidth}@{}}
\toprule
Bin length & 50 frames \\
Active clip pool & 1{,}024 clips per GPU \\
Pool refresh & rolling: 4--5 clips per iteration, full turnover every 250 iterations \\
Uniform mixing & 0.1 \\
Failure-rate cap / prior & 200$\times$ the mean rate / 1 failure in 1 attempt \\
Pre-failure credit window & 200 frames \\
Upper-body replacement probability & 0.25 (locomotion, standing, squatting and balance clips only; off for \hglobal{}) \\
Frozen reference segment probability & 0.1 \\
Low-posture weight (\hlocal{}) & $10\times$ for 1{,}753 clips, about a tenth of the pool draws: 906 floor-level clips (the
pelvis below 0.18\,m and at least 1\,s below 0.30\,m; static prone clips excepted), 427 stooping clips (a hand within
5\,cm of the floor for at least 0.1\,s with both feet down, both knees at least 0.25\,m and the pelvis at least 0.45\,m
high) and 420 low clips (the same reach with the pelvis at least 0.30\,m high, a hand down on at most half of the
frames); the takes of the floor picks of \cref{fig:ws_trackers} keep their weight \\
\bottomrule
\end{tabular}
\end{table}

\begin{table}[H]
\centering\footnotesize
\caption{\textbf{Differences of \hglobal{} from \hlocal{}}, for the motion-capture checkpoint. The last row lists
how the lidar-inertial odometry checkpoint differs.}
\label{tab:hp_global}
\begin{tabular}{@{}lp{0.58\linewidth}@{}}
\toprule
Observation & 1501-D: proprioceptive buffer 864 (9 frames $\times$ 96), dense target 335, sparse target 300, mode selector 2 \\
Tracking modes & dense, 5-point, 3-point, 2-point; sampled 1:1:1:1 (5-point evaluated in \cref{tab:main_md,tab:ht}; the released inference code runs only the dense mode) \\
Actor & transformer encoder--decoder: width 384, 6 heads, 4 layers, full self-attention over 11 tokens (5 proprioceptive frames, every second one of the buffer; 5 dense target frames; 1 sparse target); 7.6M deployed parameters \\
Control and robot model & explicit torque; wrist pitch and yaw with $k_p$ 16.8, $k_d$ 1.07, and armature 0.0043 (in evaluation, the URDF armature with these gains); other joints as for \hlocal{} \\
Rewards & world-frame anchor (1.0 / 0.5), body pose and velocity (1.0), wrist pose and velocity (1.2) and root-position (1.0) terms; anchor-relative body pose (1.0; position $\sigma = 0.05$\,m); local end-effector on the torso and wrists (1.0) and wrist orientation (1.0); survival +1.0; no wrist angular-velocity or ankle-acceleration penalty \\
Randomization & as for \hlocal{}, without the reference-pitch bias \\
Episodes & until the clip ends (no time-based timeout); extra termination at 0.5\,m world-frame body-position error \\
Pose in training & ground truth (no odometry noise); base linear velocity noise 1\,m/s; projected-gravity noise 0.01 (\hlocal{}: 0.05) \\
Motion pool & restart near failure 0.5; 1.0--1.35$\times$ time warp \\
PPO & as in \cref{tab:hp_ppo} except entropy 0.013, critic learning rate $10^{-4}$ (fixed), Adam, no minimum action std \\
Training & 65{,}536 environments (1{,}024 on each of 64 L4 GPUs, then 2{,}048 on each of 32) \\
Lidar-inertial odometry checkpoint & pose-validity input (1{,}502-D); pose and base velocity withheld in about half of the training samples; orientation-bias noise on the root and sparse targets; actor learning rate $6\times10^{-6}$ (bounds $[3\times10^{-6}, 6\times10^{-5}]$), critic $3\times10^{-5}$, KL target $9\times10^{-4}$; wrist pitch/yaw $k_p$ 8.6, $k_d$ 0.765; local end-effector weight 2.0 \\
\bottomrule
\end{tabular}
\end{table}

\section{Training Recipe Details}
\label[appendix]{supp:recipe}

This section lists the training compute of every reported checkpoint (\cref{tab:compute}) and the rewards,
robot model, and recipe details of each tracker.

\begin{table}[!h]
\centering\footnotesize
\caption{\textbf{Training compute of the reported checkpoints}, in GPU-hours on the listed GPUs. The wall time is the
median time per iteration times the number of iterations, and the GPU-hours are the wall time times the number of
GPUs. \hglobal{}
(motion capture) moves from 64 to 32 GPUs during training, with the same 65{,}536 environments.}
\label{tab:compute}
\setlength{\tabcolsep}{4pt}
\begin{tabular}{lcrrrrc}
\toprule
Recipe & GPUs & Environments & Iterations & Wall time (h) & GPU-hours & Environment steps \\
\midrule
\hlocal{} & 64 L40S & 262{,}144 & 40{,}000 & 93 & 6{,}000 & $2.5\times10^{11}$ \\
\smours{} & 1 L40S & 4{,}096 & 100{,}000 & 36 & 36 & $1.0\times10^{9}$ \\
\smisaac{} & 1 L40S & 4{,}096 & 100{,}000 & 38 & 38 & $1.0\times10^{9}$ \\
Holosoma-FastSAC & 64 L40S & 32{,}768 & 100{,}000 & 11 & 700 & $3.3\times10^{9}$ \\
\hglobal{} & 64 / 32 L4 & 65{,}536 & 35{,}000 & 149 & ${\sim}$7{,}100 & $5.5\times10^{10}$ \\
\bottomrule
\end{tabular}
\end{table}

\paragraph{Latent bound.}
Each of the 64 latent elements of \hlocal{} (\cref{sec:method:arch}) is continuous and computed as
$z = (\tanh(h + s)\,\ell - o)/w$ with the fixed constants $\ell = 15.5155$, $o = 0.5$,
$s = \operatorname{artanh}(o/\ell) \approx 0.0322$, and $w = 16$. The resulting range $(-1.00, 0.94)$ is slightly
asymmetric.

\paragraph{Rewards and terminations.}
\Cref{tab:reward} lists the reward terms of \hlocal{}.
\begin{table}[!ht]
\centering\footnotesize
\caption{\textbf{Reward terms of \hlocal{}.} All kernels are $\exp(-\lVert e\rVert^2/\sigma^2)$. The undesired-contact
term excludes the ankle-roll, wrist-yaw, and elbow links, and skips the bodies that the reference places within
2\,cm of the floor.}
\label{tab:reward}
\begin{tabular}{lcc}
\toprule
Term & $\sigma$ & Weight \\
\midrule
Anchor position (global) & 0.3\,m & 0.5 \\
Anchor orientation (global) & 0.4\,rad & 0.5 \\
Body position (anchor-relative) & 0.3\,m & 1.0 \\
Body orientation (anchor-relative) & 0.4\,rad & 1.0 \\
Body linear velocity (world frame) & 1.0\,m/s & 1.0 \\
Body angular velocity (world frame) & 3.14\,rad/s & 1.0 \\
End-effector points (torso-top, wrists) & 0.1\,m & 2.0 \\
\midrule
Action rate $\lVert a_t - a_{t-1}\rVert^2$ & -- & $-0.1$ \\
Joint-limit violation (soft limit 0.9) & -- & $-10$ \\
Undesired contacts ($>$1\,N) & -- & $-0.1$ \\
Wrist angular-velocity excess ($>$1.5\,rad/s) & -- & $-5\times10^{-3}$ \\
Ankle joint acceleration & -- & $-2.5\times10^{-6}$ \\
\bottomrule
\end{tabular}
\end{table}

An episode terminates (\cref{sec:method:reward}) when the anchor-height error exceeds 0.15\,m, or 0.75\,m when
the reference pelvis is below 0.5\,m, as in crawling or getting up, when the squared anchor-orientation error
exceeds 0.2\,rad$^2$, when the height error of any ankle or wrist exceeds 0.15\,m, relaxed in the same way, or when
the foot-position error exceeds 0.2\,m.

\paragraph{Robot model.}
\hlocal{} uses the standard model-15 G1 URDF (\cref{sec:method:dr}), whose joint limits and armature the released
robot configuration lists, with the PD gains $k_p = I\omega^2$ and
$k_d = 2\zeta I\omega$ for the armature $I$ of each joint, $\omega = 2\pi\cdot10$\,rad/s and $\zeta = 2$, and the action scale
$0.25\,\tau^{\max}/k_p$. Two motors drive each ankle and waist roll and
pitch joint through a linkage. The soft joint limits are 90\% of the range in training and 95\% in evaluation. We clip the torques at
$\tau^{\max}$ and the raw actions at $\pm20$, and set the joint friction to zero. Self-collision is enabled, except
between the pelvis and each hip-roll link, whose collision shapes overlap beyond about $83^\circ$ of hip
flexion, and which the robot model of the evaluation harness also excludes. In IsaacSim, this filter also
disables the collision between the two hip-roll links.

\paragraph{Randomization.}
Each environment draws the static and dynamic friction from $[0.3, 1.6]$ and $[0.3, 1.2]$, the restitution
from $[0, 0.5]$, base center-of-mass offsets of up to $\pm$2.5\,cm along x and $\pm$5\,cm along y and z, torso and
wrist mass scales from $[0.8, 1.5]$, and a joint calibration bias of up to $\pm0.01$\,rad. The pushes reach
linear velocities of (0.5, 0.5, 0.2)\,m/s and angular velocities of (0.52, 0.52, 0.78)\,rad/s, and there is no
action delay. The reference-pitch bias of \cref{sec:method:dr} tilts the reference orientation and the ankle
pitch observed by the actor by an angle that drifts, with a time constant of 3\,s, toward a target that we
redraw from $[-8^\circ, +3^\circ]$ every 5--20\,s, while the critic and the rewards use the true reference. Six
of every seven terrain tiles are a grid of flat 0.45\,m boxes around a flat 2\,m platform,
where the height of each box is offset uniformly by up to $\pm h$, with $h$ drawn from 1--5\,mm for each tile, and the
remaining tiles are a height field on a 0.1\,m grid, whose vertices are raised by either 0 or 20\,mm.

\paragraph{Adaptive motion sampling and augmentation.}
When a clip fails, the adaptive sampler also credits the failure to the frames before it, so that the policy
practices approaching hard segments. In addition, freezing a clip from a random frame onward, with recomputed
velocities, teaches the policy to hold a pose, and replacing the upper body of a locomotion clip with that of
another clip decouples the arm motion from the gait, as teleoperation requires (\cref{tab:hp_sampler}).

\paragraph{SuperMimic reimplementation.}
\label[appendix]{supp:sm}
\smbase{} trains the networks of \hlocal{} with the learner of SuperMimic~\cite{supermimic}, a deterministic
world model whose imagined rollouts carry soft survival weights. Besides training on flat ground and starting episodes with a failure-weighted sampler instead of the
rolling pool, it departs from \hlocal{} in three ways. Its randomization draws a single friction coefficient from $[0.3, 1.2]$ and passes the pushes to
the world model as explicit impulses. An episode ends, as in the official SuperMimic code, when the pelvis-height
error exceeds 0.25\,m or the pelvis tilts beyond the threshold of that code. The freeze-frame and upper-body
replacement augmentations are not used. It trains 4{,}096 environments, and we evaluate it after 100k iterations.

\begin{figure}[!ht]
\centering
\includegraphics[width=\textwidth]{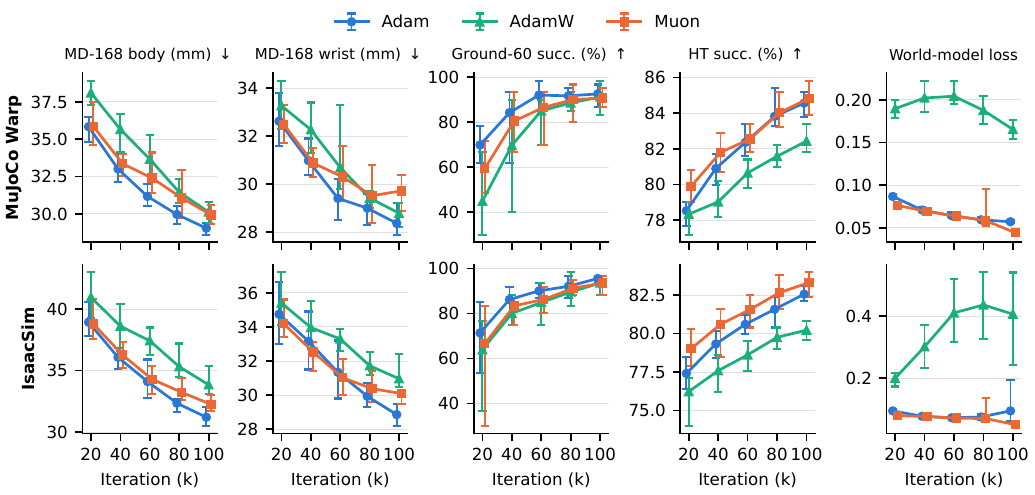}
\caption{\textbf{Optimizers for the base recipe of \smbase{}} (\cref{supp:sm}). We compare Adam without weight
decay, AdamW, and Muon in otherwise identical runs, each with ten seeds per simulator, where the Muon runs collect
data with a quarter of the environments noise-free. Points show the mean over seeds, and bars span the lowest to the
highest seed, for the MD-168 position errors and the Ground-60 and HumanTracker success, scored as in
\cref{tab:main_md,tab:ht}, and for the world-model loss averaged over the last 1k iterations. The top row shows
MuJoCo Warp, and the bottom row IsaacSim. The rows of the main tables use the full recipe with Adam.}
\label{fig:sm_optim}
\end{figure}

\paragraph{Additions for floor motions.}
\smbase{} keeps the three additions of \hlocal{} for floor motions (\cref{sec:exp:ablations}): the contact penalty
skips the bodies that the reference places within 2\,cm of the ground, the 1{,}753 low-posture clips of
\cref{tab:hp_sampler} are drawn with ten times their weight, and in IsaacSim the self-collisions between the pelvis
and the hip-roll links are filtered, as the MuJoCo model already does. Against ten seeds of the base recipe, the
additions track the bodies on MD-168 about as closely as the best seed and the wrists more closely than all of them in
MuJoCo Warp, but trail
every seed by 1--3\,mm on both in IsaacSim. The termination comparison and \cref{fig:sm_optim} use the base recipe.

\paragraph{Termination.}
The termination of \hlocal{} also bounds the end-effector and foot errors. With it, the base recipe of \smbase{}
falls on 16 of the 40 Dance-40 clips after 20k iterations in MuJoCo Warp, against 1 of 40 with the termination of
SuperMimic. After 100k iterations, it still completes only 80\% of Dance-40 in both simulators, with 9--10\,m of
root drift, although it completes more HumanTracker clips.

\paragraph{Optimizers.}
\Cref{fig:sm_optim} compares Adam, AdamW, and Muon~\cite{muon} on the base recipe, with ten seeds per optimizer and
simulator from 20k to 100k iterations. As in the official SuperMimic code, Muon updates the hidden linear weights,
and Adam the first and last layers and all biases.
\begin{itemize}[leftmargin=*,nosep]
\item \emph{Degradation.} Without changes, both MuJoCo Warp runs that we first trained with Muon degrade after 30k--40k iterations on
MotionDecode and HumanTracker, one of them also on the training-set sample of \cref{supp:data}, while the
world-model loss keeps decreasing. The first IsaacSim run does not degrade through 100k iterations.
\item \emph{Cause.} The policy comes to depend on the $\sigma = 0.1$ action noise of data collection: evaluated with
this noise, the degraded policy keeps improving, and only its deterministic rollouts degrade. As data collection
always adds the noise, training never measures the deterministic policy.
\item \emph{Fix.} Collecting data with a quarter of the environments noise-free removes the degradation. Resumed from
a degraded checkpoint at 80k iterations, the policy recovers within 5k iterations, and with ten new seeds per
simulator no run degrades through 100k iterations, with deterministic and noisy errors within about 1\,mm.
\item \emph{Result.} After 100k iterations, Muon reaches the highest HumanTracker success and a lower world-model loss
than Adam in both simulators, Adam keeps the lowest body position error, and AdamW trails both on body position and
HumanTracker success, with a higher world-model loss throughout.
\end{itemize}

\paragraph{FastSAC.}
\label[appendix]{supp:fastsac}
Holosoma-FastSAC trains the recipe of \hlocal{} with the off-policy learner of FastSAC~\cite{fastsac}, starting from
the released whole-body tracking configuration of FastSAC, which matches its paper; the released code lists every
hyperparameter. It keeps the corpus, rewards (with the reference-contact rule), terminations, observations, robot
model (without the pelvis and hip-roll self-collisions), randomization, and rolling motion pool with low-posture
clips at ten times their weight. Its single MLP actor reads the observations of the deployed \hlocal{} policy, with
zeros in the unused inputs. Three details adapt it to tracking:
\begin{itemize}[leftmargin=*,nosep]
\item the motion pool advances every 24 environment steps, the length of a PPO iteration;
\item a transition that ends with its clip bootstraps from its terminal observation, as the end of a clip is not a
failure;
\item the squashed action is bounded for each joint at its farthest joint limit, extended by $\tau^{\max}/k_p$ for
the hip, knee, and ankle-pitch joints so that the PD controller can still apply full torque at the limit, whereas
the single action scale of the released configuration would place this bound 1.16--2.19 times as far from the
default pose as the limit.
\end{itemize}
The reported model trains on 64 GPUs with 512 environments each.

\section{Motion Data Details}
\label[appendix]{supp:data}

\paragraph{Preprocessing.}
We resample every 30\,Hz source to 50\,Hz, linearly for positions and with slerp for rotations. Bones-seed
releases its G1 clips at 30\,Hz as joint rotations and root positions, so we compute their body poses by
forward kinematics with the motion loader released with SONIC~\cite{sonic}, which makes our clips match those
on which SONIC trains. We drop two helper foot-contact bodies, which leaves 30 bodies, and check the forward kinematics of the
pelvis against the root to within $2\times10^{-5}$\,m.

\paragraph{Bones-seed filter.}
The robot filter of the public SONIC training pipeline~\cite{sonic} drops, by keywords in the clip names,
motions that need props absent in simulation, such as chairs, beds, boxes, stairs, and ladders, and some
acrobatics, such as handstands and cartwheels.

\paragraph{Motion families.}
\Cref{tab:data_families} splits the corpus into motion families by keywords in the clip names. We count
100STYLE as locomotion, OMOMO as object interaction, and unmatched clips as Other, which includes about half of
the AMASS clips, whose names carry no motion label.

\begin{table}[!ht]
\centering\footnotesize
\caption{\textbf{Motion families of the training corpus} (\cref{sec:data}). Each clip belongs to the first family
that matches its name, and the hours include the mirrors. The corpus contains 149 get-up clips (0.09\%,
1.2\,h).}
\label{tab:data_families}
\begin{tabular}{lrrrr}
\toprule
Family & Clips & Share & Hours & Share \\
\midrule
Ground and floor (crawl, kneel, lie, sit, roll, get up) & 5{,}330 & 3.1\% & 24.6 & 2.9\% \\
Dance & 14{,}393 & 8.2\% & 38.5 & 4.5\% \\
Athletic and combat (jump, kick, punch, stretch) & 19{,}619 & 11.2\% & 49.4 & 5.8\% \\
Object interaction (carry, reach, place, open, wipe) & 37{,}186 & 21.3\% & 265.6 & 31.2\% \\
Gesture and expression & 22{,}664 & 13.0\% & 64.9 & 7.6\% \\
Locomotion (walk, jog, run, turn, styles) & 54{,}099 & 31.0\% & 258.9 & 30.4\% \\
Idle, posture and crouch & 13{,}399 & 7.7\% & 115.1 & 13.5\% \\
Other & 8{,}063 & 4.6\% & 34.0 & 4.0\% \\
\midrule
Total & 174{,}753 & & 851.1 & \\
\bottomrule
\end{tabular}
\end{table}

\paragraph{Training-set sample.}
To diagnose training (\cref{supp:sm}), we also score a fixed sample of 500 Bones-seed clips on which \hlocal{},
\smbase{}, and Holosoma-FastSAC train, 250 takes and their mirrors. We draw the sample deterministically and stratified by
motion category, which is given by the first token of the take name, so that it covers 84 of the 301 categories of
Bones-seed.

\paragraph{Velocity convention.}
As the sources compute velocities differently, i.e., by central differences in AMASS, 100STYLE, and OMOMO,
and by forward differences in Bones-seed, we re-derive the reference velocities of every source after forward
kinematics, as the released SONIC loader does. Specifically, we compute the joint velocities by forward
differences, the linear velocities of the bodies by central differences filtered with a Gaussian of
$\sigma = 2$ frames, and their angular velocities from the filtered relative rotation between consecutive
frames.

\section{Motion Retargeting Details}
\label[appendix]{supp:ret}

\begin{figure}[t]\centering\includegraphics[width=\linewidth]{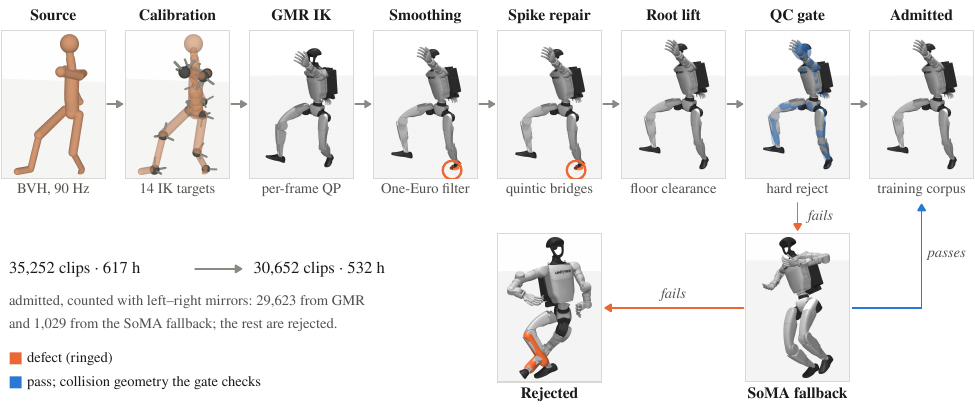}\caption{\textbf{Retargeting pipeline} on one frame of a HiPHI clip. From left to right, the panels show
the source motion, the calibrated IK targets, the GMR solution, the frame after One-Euro smoothing, after spike
repair, and after the root lift, the quality-control (QC) gate with the collision geometry in blue, and the admitted clip. In this
example, the left foot remains 48\,mm below the floor (ring) after smoothing, until the root lift. A clip that fails
the gate is solved again as a whole by SoMA, and is kept only if this solution passes.}\label{fig:ret:pipeline}\end{figure}

This section details the retargeting pipeline of \cref{sec:retarget} (\cref{fig:ret:pipeline}).

\subsection{Kinematic Retargeting}
\label[subappendix]{sec:retarget:ik}
\label[subappendix]{supp:ret:ik}
\paragraph{Source preprocessing.}
HiPHI provides 90\,Hz BVH files, which we resample to 50\,Hz, linearly for positions and with slerp for
rotations, and convert from Y-up centimeters to Z-up meters. We keep the joint rotations of HiPHI, and map its
rest skeleton onto the joint frames of the SoMA zero pose with one static rotation per joint, computed from the
limb directions in the rest pose. We use one scale for all actors, as a proportional scale for each actor reduced
the number of clips that pass every gate by about 5\% (from 31{,}528 to 29{,}832).

\paragraph{IK targets and solver.}
The 14 IK targets, i.e., the pelvis, the chest, and both thighs, shins, feet, upper arms, forearms, and hands,
use the G1 body correspondence, weights, and effector offsets of SoMA. We solve each frame with the
differential-IK quadratic program of GMR over the joint velocities (mink~\cite{mink} on MuJoCo~\cite{mujoco}),
which enforces self-collision and floor avoidance as inequality constraints on every pair of collision geometries
and between each geometry and the floor, and adds a posture task toward the previous frame. We repeat the solve of
the first frame ten times, as it has no previous frame to start from. The IK library defines its collision-velocity bound per second, while our solver steps 2\,ms, so
we rescale the bound to the solver step, which tightens it 500-fold. The agent of \cref{sec:retarget:agent} made this
change together with several others, and we do not isolate its contribution.

\paragraph{SoMA as a fallback solver.}
HiPHI provides 35{,}252 clips, i.e., 17{,}626 sequences and their left--right mirrors, and we retarget and gate each
clip separately. The GMR solution passes the QC gate on 84.0\% of these clips, and the SoMA solution, whose solver
has no self-collision objective, on 33.8\%. When the GMR solution fails and the SoMA solution passes, we keep the
SoMA solution for the whole clip, which adds 2.9 percentage points (\cref{fig:ret:steps}a).

\subsection{Smoothing and Repair}
\begin{figure*}[t]
\centering
\begin{minipage}[t]{0.48\linewidth}\centering\includegraphics[width=\linewidth]{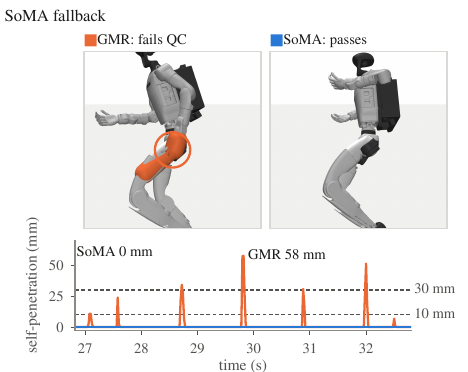}\\[-1pt]{\small (a)}\end{minipage}\hfill\begin{minipage}[t]{0.48\linewidth}\centering\includegraphics[width=\linewidth]{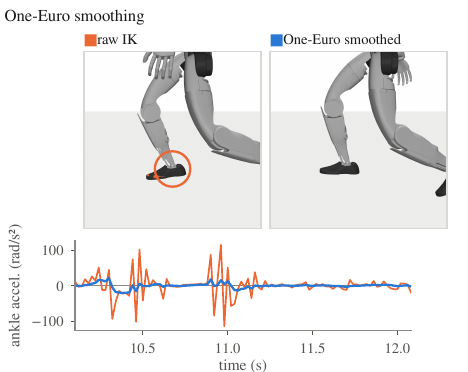}\\[-1pt]{\small (b)}\end{minipage}\\[4pt]
\begin{minipage}[t]{0.48\linewidth}\centering\includegraphics[width=\linewidth]{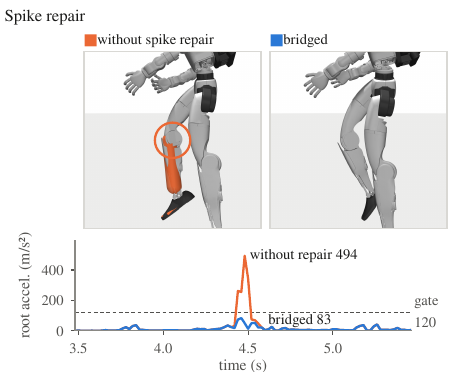}\\[-1pt]{\small (c)}\end{minipage}\hfill\begin{minipage}[t]{0.48\linewidth}\centering\includegraphics[width=\linewidth]{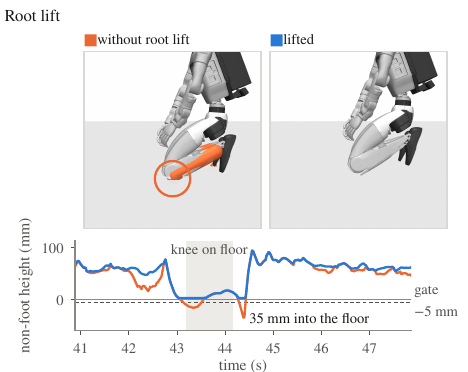}\\[-1pt]{\small (d)}\end{minipage}
\caption{\textbf{When each stage is needed.} HiPHI clips without (orange) and with (blue) each stage, at the frame of
the largest defect, together with the signal that the QC gate checks. (a)~The GMR solution drives the left hand
58\,mm into the thigh, while the SoMA solution passes. (b)~Smoothing removes a tremor of the right ankle.
(c)~Spike repair removes a spike at the right knee, lowering
the root acceleration from 494 to 83\,m/s$^2$ (gate 120\,m/s$^2$). (d)~Without the root lift, a kneeling shin sinks
35\,mm into the floor (gate 5\,mm). The shaded interval marks where the human knee is on the floor.}
\label{fig:ret:steps}
\end{figure*}

\label[subappendix]{sec:retarget:repair}
As the IK solves one frame at a time and does not know which bodies of the human are in contact, we smooth,
repair, and lift its solution before the QC gate (\cref{fig:ret:steps}).

\paragraph{Smoothing.}
A causal One-Euro filter~\cite{oneeuro} on the 29 joint angles, with a minimum cutoff of 1.2\,Hz, $\beta = 0.4$,
and a derivative cutoff of 1.0\,Hz, removes the frame-to-frame tremor of the IK (\cref{fig:ret:steps}b).
However, smoothing also softens the per-frame collision and contact corrections of the IK. Of
the 31{,}528 GMR clips that pass every gate without smoothing, 3{,}006 (9.5\%) fail after smoothing, about half of
them through self-collision and most others through a floating foot, while only 161 failing clips pass. Together
with 940 clips whose processing failed only in the run without smoothing, 29{,}623 smoothed GMR clips pass every
gate. We reject the clips that fail after smoothing, instead of reverting them to their solution without smoothing.

\paragraph{Spike repair.}
We bridge each isolated kinematic spike, such as a single-frame jump in joint velocity, with a quintic segment
(\cref{fig:ret:steps}c), which we widen by up to 12 frames on each side of the spike until every kinematic gate
passes. We reject a clip if the repair rewrites more than 10\% of its frames.

\paragraph{Root lift.}
We raise the root until the lowest collision geometry clears the floor by 3\,mm (\cref{fig:ret:steps}d). The lift
keeps the feet, knees, hands, pelvis, and chest on the floor wherever they are on the floor in the source, and is
limited to 25\,cm, 3\,m/s, and 150\,m/s$^2$. We reject the clips that need more.

\subsection{Quality-Control Gates}
\label[subappendix]{sec:retarget:qc}
We convert every output to the training format with MuJoCo forward kinematics and float32 values. After structural
checks, i.e., finite values, unit quaternions, the data layout, a frame rate of 50\,Hz, and the joint limits, every
clip must pass the gates of \cref{tab:ret:gates}, which we evaluate on the stored values, i.e., on the exact data that
the tracker trains on. Where the source shows hand contact with the body, such as hands on the thighs, we relax only the
\emph{persistent} self-collision gate, while the 30\,mm cap on the penetration depth still applies.

\begin{table}[t]
\centering\footnotesize
\caption{\textbf{Quality-control gates.} A clip is rejected if it exceeds any of these thresholds. We report the pose
error with respect to the source, but do not gate it.}
\label{tab:ret:gates}
\begin{tabular}{ll}
\toprule
Gate & Threshold \\
\midrule
Floor penetration (non-foot / foot) & $>$5\,mm / foot $z < -5$\,mm \\
Self-collision depth & $>$30\,mm on any frame \\
Persistent self-collision & $>$10\,mm on $>$1\% of frames \\
Floating while source in contact & $>$50\,mm on $>$5\% of contact frames, or $>$5 consecutive \\
Root speed (lin / ang) & 12\,m/s / 20\,rad/s \\
Root acceleration (lin / ang) & 120\,m/s$^2$ / 250\,rad/s$^2$ \\
Joint speed / acceleration & 25\,rad/s / 500\,rad/s$^2$ \\
Acceleration sign-flip fraction (tremor) & $>$0.25 \\
Repaired-frame fraction & $>$10\% \\
\bottomrule
\end{tabular}
\end{table}

\subsection{Agentic Iteration}
\label[subappendix]{sec:retarget:agent}
\label[subappendix]{supp:ret:agent}
An LLM coding agent automatically iterated on this recipe, raising the number of HiPHI clips that pass every
gate from 27 with the initial pipeline to 31{,}528 before smoothing.

\paragraph{Iterations.}
In each iteration, the agent inspects which gates each clip fails, how these failures distribute over motion types
and over the contacts of the source, and videos of the source and the retargeted motion side by side. It then
changes one part of the pipeline, i.e., a parameter or a repair stage, reruns the pipeline on cloud CPU workers,
and keeps the change only if more clips pass all gates, not only the gate that the change targets. Each change is
first tested on a small subset of clips, and then on a larger, fixed subset, on which it must not lower the number
of passing clips, before it runs on the full corpus.

\paragraph{Human oversight.}
We fixed the gates, which the agent could not edit, and the pipeline stores, for every clip, the settings that
produced it. In addition, we set the overall criteria, i.e., to tolerate pose errors with respect to the source but
to enforce the physical gates strictly, approved each version of the corpus, reviewed videos, and made the main
design decisions, such as no longer using SoMA as the primary solver. We also asked the agent to test alternatives,
such as a cap on the joint velocity, which made the results worse. The agent found the rescaled collision bound,
the spike repair, and the One-Euro smoothing stage. It also added an option
that keeps a repair or smoothing stage for a clip only if the stage adds no failing gate, which we do not use for
smoothing (\cref{sec:retarget:repair}).

\subsection{Resulting Corpus and Physical Feasibility}
\label[subappendix]{sec:retarget:corpus}
The resulting HiPHI corpus holds 30{,}652 clips, of which 29{,}623 come from GMR and 1{,}029 from the SoMA
fallback. As each clip and its left--right mirror pass QC separately, these are 15{,}270 source clips and 15{,}382
mirrors. The released pipeline takes 5--7\,ms of one CPU core per 50\,Hz frame, or about 190 CPU-hours for all of HiPHI.

The QC gates check the physical validity of the frames, but not whether the motion can be performed on flat ground
without props. For this, three clip-level rules flag a clip when the pelvis stays more than 1.2\,m above the floor
for over 0.5\,s, when the clip starts with no body near the floor, or when no geometry or fingertip is within 5\,cm
of the floor for over 0.5\,s outside a ballistic arc. These rules flag 0.4\% of the public training clips that we
scanned, while our reimplementation of the filter of SuperMimic~\cite{supermimic}, which combines the keyword list of
SONIC with forward-kinematics checks for elevated support, ground penetration, and floating feet, flags 2.1\%. A fourth rule, which flags floor
penetrations deeper than 3\,cm, mostly detects a different convention for the ground offset rather than infeasible
motion, as the AMASS and 100STYLE clips sit about 3\,cm into the floor.

\section{HumanTracker Protocol}
\label[appendix]{supp:ht}
The SONIC releases and Humanoid-GPT run through the adapters provided by the benchmark, and ScaleBFM,
HoloMotion, HEFT, and TeleopIT through the policy code that they provide for MotionDecode. For SONIC v1.1 and
Humanoid-GPT, we reproduce their published results to within 0.3 and 0.6 points of success per motion family. Clips with an
undefined joint error only count toward the success rate.
The robot model of HumanTracker (\cref{sec:eval}) has an armature of 0.01 and a torque limit of 13.4\,Nm for
the wrist pitch and yaw joints, a joint friction of 0.1, and a physics rate of 1\,kHz, while SONIC keeps its own
torque limits in this benchmark, i.e., 25\,Nm for the ankle and waist roll and pitch joints and 5\,Nm for the
wrists.

\section{Ablations}
\label[appendix]{supp:ablations}
\label[appendix]{sec:exp:ablations}

We train variants of a base recipe that each change one component and compare them after 40k iterations
(\cref{tab:ablate}). The base recipe is \hlocal{} without its three additions for floor motions.

\paragraph{Summary.}
Against a periodic re-draw, which replaces all 1{,}024 clips of a GPU every 250 iterations and resets every
environment, the rolling pool refresh lowers the body and wrist errors and raises the HumanTracker success, and so
does the $4\times$ larger batch. The scaled $\tanh$ bound of the latent (\cref{sec:method:arch}) also tracks more
closely than a plain $\tanh$, and a pelvis anchor beats a torso anchor on all three metrics. Repairing small
floor offsets and joint-limit excursions of the corpus clips, a step separate from our retargeting, raises the
tracking errors, and so does a penalty on joint targets near their limits. Exempting the contacts that the
reference also makes from the contact penalty costs a little accuracy, but serves floor motions: together with the
filter of the self-collisions between the pelvis and the hip-roll links and the ten-fold low-posture weight, it
forms the three additions of the final recipe, with which \hlocal{} completes 78.3\% of the Ground-60 clips
(\cref{tab:main_md}). A small transformer command encoder tracks the wrists slightly more closely than the MLP
but the body less closely and completes fewer HumanTracker clips; both encoders run well within the 20\,ms
control period, and \hlocal{} keeps the MLP.

\begin{figure}[H]
\centering
\includegraphics[width=0.95\linewidth]{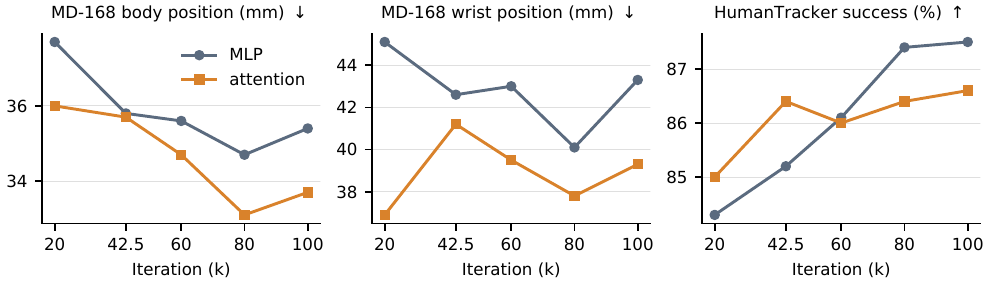}
\caption{\textbf{MLP versus attention command encoder} with the development recipe. We report the MD-168 body and
wrist position errors (mm) and the HumanTracker success (\%) at the same checkpoints of both runs.}
\label{fig:encoder_trend}
\end{figure}

\paragraph{Runs.}
All rows of \cref{tab:ablate} use this base recipe at 40k iterations, except for the
rows marked as \emph{development}, which use earlier recipes and an earlier version of the robot model. Of these,
the assistive-wrench rows use the development recipe, which trains 65{,}536 environments, at 50k iterations, and the precision rows
use an earlier version of the development recipe at 35.5k iterations. HumanTracker clips without a valid joint
error count as failures here and in \cref{fig:encoder_trend}.

\begin{table}[t]
\centering\footnotesize
\caption{\textbf{Single-variable ablations of the base recipe.} MD-168 pools Locomotion-80, Manipulation-48, and
Dance-40, excluding Ground-60 as too noisy (\cref{supp:noise}), and HT is the HumanTracker test success. In the
rows marked with $^\ast$, clips without a valid joint error count as failures (35 in the assist-none row, 43 in the
torso row, and 214 in the assist-curriculum row), and 129 in the row marked with $^\diamond$. Bold marks the better value of each
pair.}
\label{tab:ablate}
\begin{tabular}{lccc}
\toprule
& \multicolumn{2}{c}{MD-168 (mm)} & HT \\
\cmidrule(lr){2-3}\cmidrule(lr){4-4}
Variant & Body pos & Wrist pos & Succ.\ (\%) \\
\midrule
Pool refresh: periodic re-draw & 33.6 & 38.3 & 89.6 \\
Pool refresh: rolling & \textbf{32.0} & \textbf{36.3} & \textbf{90.7} \\
\midrule
Batch: 65{,}536 envs & 35.1 & 39.7 & 87.3 \\
Batch: 262{,}144 envs & \textbf{32.0} & \textbf{36.3} & \textbf{90.7} \\
\midrule
Corpus: as built & \textbf{32.0} & \textbf{36.3} & \textbf{90.7} \\
Corpus: repaired & 33.1 & 37.6 & 90.2 \\
\midrule
Contact penalty: every undesired contact & \textbf{32.0} & \textbf{36.3} & \textbf{90.7} \\
Contact penalty: contacts the reference also makes exempt & 32.6 & 36.9 & 90.2 \\
\midrule
Latent bound: scaled $\tanh$ & \textbf{32.0} & \textbf{36.3} & \textbf{90.7} \\
Latent bound: plain $\tanh$ & 32.7 & 37.1 & 90.3 \\
\midrule
Target penalty: none & \textbf{32.0} & \textbf{36.3} & \textbf{90.7} \\
Target penalty: targets beyond 95\% of the joint range & 38.7 & 39.9 & 88.0 \\
\midrule
Anchor body: pelvis & \textbf{32.0} & \textbf{36.3} & \textbf{90.7} \\
Anchor body: torso & 33.1 & 38.6 & 88.2$^\ast$ \\
\midrule
Assistive root wrench (development): none & \textbf{34.3} & \textbf{40.4} & \textbf{84.9}$^\ast$ \\
Assistive root wrench (development): curriculum & 35.8 & 42.6 & 84.6$^\ast$ \\
\midrule
Encoder: MLP & \textbf{32.0} & 36.3 & \textbf{90.7} \\
Encoder: attention & 33.1 & \textbf{35.6} & 90.0 \\
\midrule
Training precision (development): fp32 & \textbf{37.0} & \textbf{43.3} & \textbf{83.6} \\
Training precision (development): bf16 autocast & 39.6 & 45.1 & 81.2$^\diamond$ \\
\bottomrule
\end{tabular}
\end{table}

\paragraph{Corpus repair.}
To build the repaired corpus, we shift every clip whose planted feet are more than
10\,mm above or below the floor height of Bones-seed onto that floor, smoothly lift the root by at most 0.25\,m where
a floor check still fails, and clamp joint-limit excursions of up to 0.01\,rad. We then recompute the body poses by
forward kinematics, re-check every repaired clip, and drop it if any check fails, which repairs 29{,}910 clips and
drops 334. Unlike the retargeting repairs of \cref{supp:ret}, which decide which HiPHI clips enter the corpus, this
repair edits clips of every source, such as the AMASS and 100STYLE clips with the floor offset of
\cref{sec:retarget:corpus}. The repaired corpus loses mostly on Locomotion and Dance and is level on Manipulation,
so that, with one run per variant, we only conclude that the repair does not help. These floor defects also do not
prevent a tracker from learning ground motion, as the SuperMimic recipe completes nearly every Ground-60 clip on the
same unrepaired corpus (\cref{tab:main_md}). We have not tested why the repair does not help, but one possible
explanation is that the tracker already tolerates floor offsets of a few centimeters, as it trains on uneven ground,
so that the repair changes the references without removing a source of error.

\paragraph{Latent bound.}
The latent ablation replaces the scaled $\tanh$ bound of \hlocal{} (\cref{sec:method:arch}) with a plain $\tanh$ bound,
$z = \tanh(h)$, with the range $(-1, 1)$, and neither variant quantizes the latent. After 40k iterations,
the plain $\tanh$ bound trails in body and wrist position by more than the
checkpoint-to-checkpoint variation. On the development recipe, it also trails on every metric, as the scaled $\tanh$ bound
leads by 1.6\,mm in body position (95\% interval 0.9--2.4\,mm), 2.8\,mm in wrist position, and 2.0 points of
HumanTracker success, paired over five checkpoints (40.5k--50k iterations).

\paragraph{Joint-range target penalty.}
The joint-range target penalty of \cref{sec:exp:deploy} also shortens the training episodes, i.e., the policy
fails earlier on its training clips, so its loss at 40k iterations (\cref{tab:ablate}) is not an artifact of the
evaluation.

\paragraph{Anchor body.}
Every reported model anchors its command and rewards at the pelvis, in whose frame the harness also scores
them. The torso anchor is level on Manipulation but 1.0 and 2.7\,mm behind in body position on
Locomotion and Dance, where the torso moves more relative to the pelvis. This gap is an upper bound, since under
a torso anchor the torso point of the end-effector reward always has zero error, which dilutes that term by
about a third.

\paragraph{Assistive root wrench.}
On the development recipe, we also test an assistive wrench on the pelvis toward the reference root pose, as in
ZEST~\cite{zest}. The wrench consists of a PD term on the pelvis pose error and a feedforward term, is scaled for
each clip by a shrinkage estimate of its failure rate, and is capped at 1.5 body weights. It is applied during the
first 40\% of training and annealed to zero by 70\%, 5\% of the environments never receive it, and only the critic
observes it. The wrench does not help, as, after 50k iterations, it trails by 1.3--1.7\,mm in body position on
Locomotion, Manipulation, and Dance, and ties on HumanTracker.

\paragraph{Training precision.}
Training with bf16 autocast raises the throughput by 5\%, but loses on every metric compared with fp32
(\cref{tab:ablate}).

\paragraph{Command encoder.}
\label[appendix]{supp:encoder}
The attention encoder is a small transformer with a width of 128, 4 heads, and 2 layers. It uses one token for
the proprioceptive history and one for each future frame, with rotary position embeddings at the 50\,Hz frame rate,
and maps them to the same bounded latent through a learned-query readout. A forward pass with a batch size of 1 on
one CPU thread of an Apple M-series laptop takes 0.87\,ms for the MLP model (12.8M parameters) and 0.68\,ms for the
attention model (9.7M). With the development recipe (\cref{fig:encoder_trend}), the attention encoder leads in body
and wrist position at every checkpoint, unlike with the base recipe at 40k iterations,
but its run takes 28\% more GPU time (10.4 instead of 8.1\,s per iteration).

\section{Workspace Coverage}
\label[appendix]{supp:workspace}

The training corpus covers most of the evaluation workspace. Relative to the pelvis, about 90\% of the MD-168
wrist and ankle regions lie inside the region that the corpus covers (89\% and 91\%, and 97\% for the wrists during
stationary manipulation). The held-out errors of \hlocal{} therefore mostly measure generalization to new motions in
familiar hand and foot positions, not to new regions of the workspace.

\Cref{fig:ws_trackers} shows where each tracker loses accuracy. Its errors are weighted by frame, and are thus higher than
the clip-weighted means of \cref{tab:main_md}, as the long Dance-40 clips hold about half of the MD-168 frames. The
region where all five trackers are most accurate (\cref{sec:exp:main}) covers the wrists at mid height and moderate
reach and the ankles close to the pelvis. Low wrist targets are the hardest for most local trackers, and the highest
ones for \hglobal{}. In the world frame, the roots of the local trackers drift by up to about a meter.

\section{Deployment Behavior}
\label[appendix]{sec:exp:deploy}

Body errors in the pelvis frame do not show whether the robot keeps up with the reference, so we also measure
walking speed and torque demand (\cref{supp:eval,tab:deploy}). \hlocal{} walks
more slowly than SONIC, without falling. SONIC exceeds the 35\,Nm limit of the ankle-pitch joints in PD torque
demand, only when it commands joint targets outside the joint range, and \hlocal{} nearly reaches it; its rare
steps above the limit also all have targets outside the range. Penalizing targets beyond 95\%
of the joint range in the base recipe removes them and lowers the torque demand to 69\% of the limit, but worsens
tracking on every held-out suite (\cref{tab:ablate,supp:ablations}).

\begin{table}[!ht]
\centering\footnotesize
\caption{\textbf{Deployment behavior} on the 12 walking clips of 45\,s. We report the walking speed, the
ankle-pitch PD torque demand relative to the 35\,Nm limit, and the share of control steps with ankle- or waist-pitch
targets outside the joint range.}
\label{tab:deploy}
\begin{tabular}{lccc}
\toprule
 & Speed (m/s) & Torque demand (\%) & Targets out of range (\%) \\
\midrule
SONIC~\cite{sonic} & 0.266 & 123 & 11.5 \\
\oursrow \hlocal{} & 0.242 & 98 & 12.0 \\
\bottomrule
\end{tabular}
\end{table}

\section{Payload Robustness}
\label[appendix]{supp:payload}
\label[appendix]{sec:exp:payload}

The payload study repeats Locomotion-80 and Manipulation-48 under five configurations: our teleoperation
payload (\cref{fig:payload_configs}), the same with the backpack 5\,cm further back, 0.5 or 1.0\,kg on each
wrist, and a $1.5\times$ heavier backpack (\cref{tab:payload}). Under these payloads, only HEFT and HoloMotion change their
Locomotion root error by more than 0.06\,m. The wrist error even \emph{decreases} for six of the eight trackers,
which does not indicate robustness. In fact, on the nominal robot, their wrists sit 8--25\,mm above the reference
relative to the pelvis, and the added hand mass pulls them down by 6--23\,mm. The wrist error only rises for the
two ScaleBFM models, whose wrists sit 0--6\,mm above the reference. Finally, the 1.0\,kg wrist load affects the
ScaleBFM models and \hglobal{} the most.

\begin{table}[H]
\centering\footnotesize
\caption{\textbf{Payload robustness on MotionDecode.} We report the nominal root error and translation gain on
Locomotion-80 and their change under the configuration of \cref{fig:payload_configs}(b) and with its backpack 5\,cm
further back, the change of the wrist error on Manipulation-48 under the same two configurations, and the change in
progress (points) and wrist error (mm) in the mass sweep. $^\S$See \cref{tab:main_md}.}
\label{tab:payload}
\setlength{\tabcolsep}{4pt}
\begin{tabular}{lcccccccccc}
\toprule
& \multicolumn{3}{c}{Root error (m)} & \multicolumn{3}{c}{Transl.\ gain ($\times100$)}
& \multicolumn{2}{c}{$\Delta$ wrist (mm)} & \multicolumn{2}{c}{Mass sweep} \\
\cmidrule(lr){2-4}\cmidrule(lr){5-7}\cmidrule(lr){8-9}\cmidrule(lr){10-11}
 & & $\Delta$ & $\Delta$ & & $\Delta$ & $\Delta$ & & & Wrist & Backpack \\
Model & Nom. & payload & $-5$\,cm & Nom. & payload & $-5$\,cm
& payload & $-5$\,cm & $+1.0$\,kg & $\times1.5$ \\
\midrule
SONIC~\cite{sonic} & 0.39 & +0.01 & +0.03 & 99.9 & $-$2.1 & $-$2.4 & $-$10.9 & $-$10.1 & $-$0.4 / $-$10.7 & +0.0 / $-$10.5 \\
SONIC v1.1~\cite{sonic} & 0.54 & +0.01 & +0.01 & 94.3 & $-$0.4 & $-$0.5 & $-$5.6 & $-$3.8 & +0.0 / $-$9.6 & +0.0 / $-$4.3 \\
HEFT~\cite{heft} & 0.45 & +0.07 & +0.11 & 93.6 & $-$1.2 & $-$2.1 & $-$5.8 & $-$4.3 & $-$1.2 / +6.8 & +0.0 / $-$4.5 \\
ScaleBFM-M~\cite{scalebfm} & 0.96 & $-$0.05 & $-$0.01 & 84.1 & +0.2 & +0.0 & +5.7 & +6.0 & +0.0 / +30.2 & +0.0 / +8.2 \\
ScaleBFM-XL~\cite{scalebfm} & 0.65 & +0.02 & +0.06 & 90.2 & $-$0.5 & $-$1.2 & +8.7 & +10.5 & $-$1.4 / +29.4 & +0.0 / +11.9 \\
HoloMotion~\cite{holomotion}$^\S$ & 0.90 & +0.17 & +0.20 & 89.2 & $-$2.4 & $-$3.1 & $-$2.4 & $-$1.8 & +0.0 / +14.7 & +0.0 / $-$1.0 \\
\oursrow \hlocal{} & 0.59 & $-$0.02 & +0.02 & 95.8 & $-$2.0 & $-$2.8 & $-$20.4 & $-$18.5 & +0.0 / +5.2 & +0.0 / $-$19.2 \\
\oursrow \hglobal{} & 0.06 & +0.00 & +0.00 & 99.9 & +0.0 & +0.0 & $-$4.2 & $-$4.7 & $-$1.4 / +19.3 & +0.0 / $-$4.0 \\
\bottomrule
\end{tabular}
\end{table}

\section{Gait Study}
\label[appendix]{sec:exp:gait}

\paragraph{Protocol.}
On hardware, we observed that the backward translation gain of SONIC is 0.48--0.69 of its forward gain, a
direction-specific behavior that aggregate metrics do not reveal. In \emph{open loop}, we replay
70 windows, i.e., 22 LAFAN1~\cite{lafan} windows of forward, backward, sideways, and turning walks at 0.75, 1.0,
and 1.25 times their speed, and four start-and-stop windows. In \emph{closed loop}, a virtual operator loops a
robot-relative walking window as the reference, as in teleoperation, until the robot has moved 1.5\,m or turned
90$^\circ$, and then blends to rest. SONIC v1.1 receives its command in the layout that it was trained
with, and ScaleBFM-M and HoloMotion, whose commands contain the motion of the reference root, receive the
commanded root path as the root of their reference. The sideways results are based on only five windows (2 left, 3 right),
and we exclude one floor-roll window on which every model falls. We do not evaluate global trackers, including
\hglobal{}, in their root-pose mode.

\begin{figure}[H]
\centering
\includegraphics[width=\textwidth]{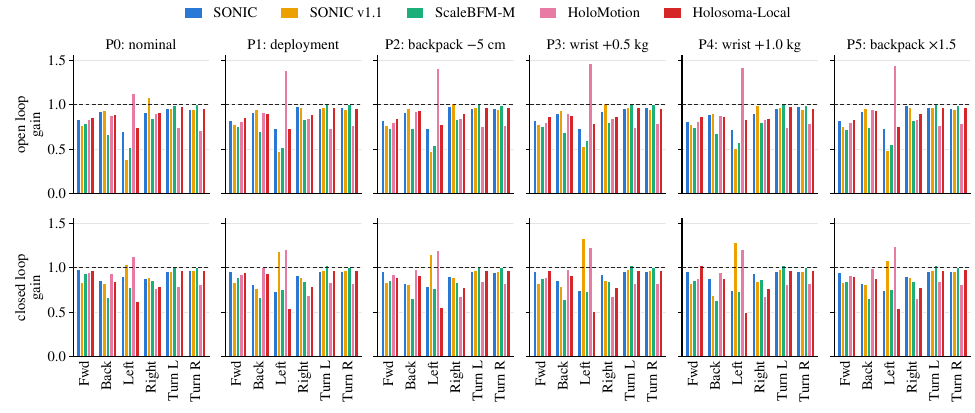}
\caption{\textbf{Translation gain by direction and robot configuration} (turn gain for turns), in open loop (top)
and in closed loop with a virtual operator (bottom). The configurations are the nominal robot (P0), the
deployment configuration with backpack, cameras, and grippers (P1), the backpack moved by $-5$\,cm (P2),
$+0.5$ and $+1.0$\,kg on each wrist (P3--P4), and a $1.5\times$ heavier backpack (P5), and the dashed line marks a
gain of one.}
\label{fig:gait}
\end{figure}

\paragraph{Results.}
In closed loop (\cref{fig:gait}), \hlocal{} covers about as much of a forward walk as SONIC and more
than SONIC v1.1, ScaleBFM-M, and HoloMotion, and about as much of a backward walk as the SONIC releases, but only
about two thirds of a leftward walk, less than any baseline. HoloMotion covers more of a backward walk, and
ScaleBFM-M about two thirds of it, although both follow the commanded root path. The backward-to-forward ratio of
\hlocal{} is close to that of SONIC, whose backward shortfall on hardware the study reproduces only in a weaker
form. The leftward shortfall of \hlocal{} does not appear in open loop, and while it steps left, its hip roll
torque never saturates and its hip pitch torque only rarely. HoloMotion turns less than commanded in both
settings, and only SONIC v1.1 falls. On the hardware reference clips replayed in simulation, \hlocal{} follows
forward motion with a gain of 0.73, compared with 0.5--0.6 for the baselines.

\section{Teleoperation Task Details}
\begin{figure*}[p]
\centering
\begin{subfigure}{\textwidth}
\centering
\includegraphics[width=0.95\linewidth]{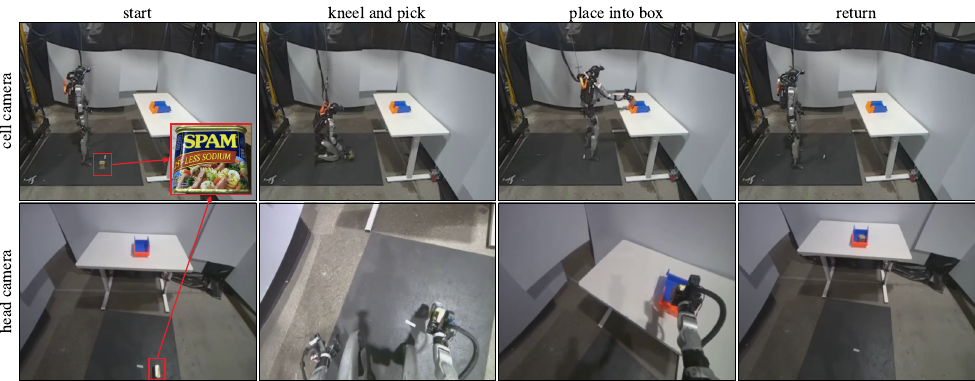}
\caption{Kneel-down PnP. The robot kneels, picks a SPAM can from the floor, places it into the box, and
returns to its start pose.}
\label{fig:teleop_kneel}
\end{subfigure}
\par\medskip
\begin{subfigure}{\textwidth}
\centering
\includegraphics[width=0.95\linewidth]{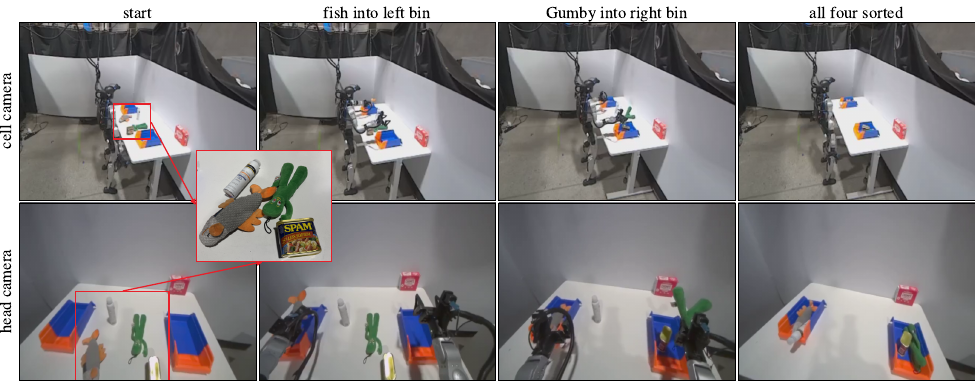}
\caption{Bi-manual PnP. The left hand places a fish toy and a bottle into the left bin, and the right hand
places a SPAM can and a Gumby figure into the right bin.}
\label{fig:teleop_bimanual}
\end{subfigure}
\par\medskip
\begin{subfigure}{\textwidth}
\centering
\includegraphics[width=0.95\linewidth]{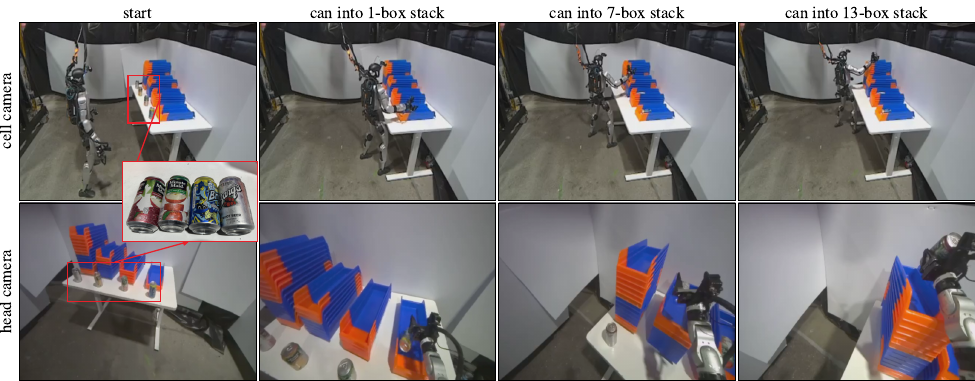}
\caption{4-in-a-row PnP. The robot places each can into the top box of the stack behind it, from the
shortest stack to the tallest.}
\label{fig:teleop_fourrow}
\end{subfigure}
\caption{\textbf{The three teleoperation tasks}, one episode each, from a cell camera (top) and the head camera of the robot (bottom). Insets show the objects, whose placement and order are randomized.}
\label{fig:teleop_tasks}
\end{figure*}

\label[appendix]{supp:teleop}
In the study of \cref{sec:exp:teleop}, data collectors (DCs) run the trials and annotators (DAs)
review them. \Cref{fig:teleop_tasks} shows one episode of each task.

\paragraph{Data collection.}
Each trial involves two DCs, i.e., an operator, who teleoperates the robot, and a resetter, who sets up a new
randomization and confirms that the cell is clear. An episode runs from the standing start pose, with the arms
down, to the return to this pose after the last placement, with the arms lowered over 1--2\,s, and the operator
marks its start and end with the handheld controller. Until the recording ends, nobody touches the objects,
bins, or boxes, which the robot may bump. The robot may re-grasp an object after a miss or slip, but a grasped
object must stay visible in the head camera, and objects may be released about 10\,cm above their target and
need not land upright.

\paragraph{Failures and annotation.}
A trial fails if the robot falls, stumbles, is caught by the safety gantry, or ends in an awkward arm
configuration, such as a rolled shoulder, even if it recovers. A trial also fails if anyone touches or re-places
an object, if an object falls off the table or ends outside its target, if the task is incomplete, or if one of
the task-specific failures below occurs. In all these cases, the operator ends the episode on the spot and records it as a failure. A trial
with a grasp attempt longer than 8\,s from the first reach to the closed gripper also counts as a failure. The DAs
review the recordings, may move the start and end frames, and apply the same conditions. In \cref{tab:teleop},
a start that the operator ends within 5\,s is considered accidental and ignored.

\subsection{Kneel-down PnP}
The robot starts on a floor mark facing a table, with its yaw varied within $\pm$15$^\circ$ and the can in
view. A SPAM can, with a random label orientation, stands on the floor 40--60\,cm ahead and slightly right of
center, varied by $\pm$15\,cm laterally within reach of a kneel on the right knee, and a box stands at a varied
spot near the front edge of the table, among distractors. The operator kneels on the right knee in 2--3\,s, with
the left foot half a step forward and the torso upright, grasps the can with the right gripper, stands up in
2--3\,s while holding it at the chest, walks to the table, releases the can into the box, and walks backward to
the mark. The trial also fails if a phase is missing, if the can tips before the grasp or is dropped, or if the
robot drops or bounces onto its knee, kneels on both knees, sits back on its heel, puts a hand on the floor, or
touches the table with its body, while touching the box is allowed.

\subsection{Bi-manual PnP}
The robot starts at a fixed spot at the near corner of a table of fixed height, facing along it, with identical
bins fixed at its front left and front right. A SPAM can and a Gumby plush lie on the right, and a fish plush and a
bottle on the left, each varied by $\pm$10--15\,cm on its side, with a random label or toy orientation, and
reachable without walking, although a small step is allowed. In any order, each hand lifts the two objects on its
side by about 10\,cm and places them into its bin, with the bottle upright, while the idle arm stays still with its
gripper open. The arms never cross, and a hand only moves an object within reach of the other hand if needed. The
trial also fails if an object ends in the wrong bin or if the arms or grippers collide, while dropping an object
onto the table is allowed.

\subsection{4-in-a-row PnP}
The robot starts at the same table corner. Four stacks of identical boxes, with heights of 13, 7, 4, and 1
boxes from left to right, stand at fixed spots along the table, and each target is the top box of a stack. A
juice or tea can, never carbonated, stands in front of each stack, with the assignment shuffled in each trial, its
position varied by $\pm$5--10\,cm, and a random label orientation. Using the right hand, with the left arm still,
the operator lifts each can by about 10\,cm and drops it into the top box of its stack from just above the rim, in
the fixed order of the 1-, 4-, 7-, and 13-box stacks, lifting it straight up first for the two tall stacks. A small
step toward a stack out of reach is allowed. The trial also fails if a can ends in the wrong stack, if the stacks
are filled out of order, or if a stack tips or a box falls.

\end{document}